\documentclass[pdflatex,sn-basic]{sn-jnl}

\usepackage{graphicx}%
\usepackage{multirow}%
\usepackage{amsmath,amssymb,amsfonts}%
\usepackage{amsthm}%
\usepackage{mathrsfs}%
\usepackage[title]{appendix}%
\usepackage{textcomp}%
\usepackage{manyfoot}%
\usepackage{booktabs}%
\usepackage{algorithmicx}%
\usepackage{algpseudocode}%
\usepackage{listings}%

\usepackage{comment}%
\usepackage{caption}%
\usepackage{CJKutf8}%
\usepackage[table, dvipsnames]{xcolor}%
\usepackage{longtable}%
\usepackage{csquotes}%
\usepackage{lmodern}%
\usepackage{pgfplots}%
\usetikzlibrary{patterns}%
\usepackage{array}%

\theoremstyle{thmstyleone}%
\theoremstyle{thmstyletwo}%

\theoremstyle{thmstylethree}%

\newsavebox{\refbox}
\sbox{\refbox}{ABCDEFGHIJKLMNOPQRSTUVWXYZabcdefghijklmnopqrstuvwxyzÄÖÜäöüßgpqy}
\newcommand{\redhighlight}[2]{%
  \begingroup
  \setlength{\fboxsep}{0pt}%
  \colorbox{#1}{\raisebox{0pt}[\ht\refbox][\dp\refbox]{#2}}%
  \endgroup
}

\newcommand{\gcolor}[1]{\cellcolor{ForestGreen!#1!White}}

\begin{document}

\title[When the Coffee Feature Activates on Coffins]{When the Coffee Feature Activates on Coffins:\newline An Analysis of Feature Extraction and Steering for Mechanistic Interpretability}

\author*[1]{\fnm{Raphael} \sur{Ronge}}\email{raphael.ronge@hfph.de}
\equalcont{These authors contributed equally to this work.}

\author[1]{\fnm{Markus} \sur{Maier}}\email{markus.maier@hfph.de}
\equalcont{These authors contributed equally to this work.}

\author[2]{\fnm{Frederick} \sur{Eberhardt}}\email{fde@caltech.edu}

\affil[1]{\orgdiv{Department of Philosophy of Nature and Technology}, \orgname{Munich School of Philosophy}, \orgaddress{\street{Kaulbachstra\ss e 31a}, \city{Munich}, \postcode{80539}, \state{Bavaria}, \country{Germany}}}

\affil[2]{\orgdiv{Division of the Humanities and Social Sciences}, \orgname{California Institute of Technology}, \orgaddress{\street{1200 East California Boulevard}, \city{Pasadena}, \postcode{91125}, \state{CA}, \country{USA}}}


\abstract{Recent work by Anthropic on Mechanistic interpretability claims to understand and control Large Language Models by extracting human-interpretable features from their neural activation patterns using sparse autoencoders (SAEs). If successful, this approach offers one of the most promising routes for human oversight in AI safety.  We conduct an initial stress-test of these claims by replicating their main results with open-source SAEs for Llama 3.1.
While we successfully reproduce basic feature extraction and steering capabilities, our investigation suggests that major caution is warranted regarding the generalizability of these claims.
We find that feature steering exhibits substantial fragility, with sensitivity to layer selection, steering magnitude, and context. We observe non-standard activation behavior and demonstrate the difficulty to distinguish thematically similar features from one another.
While SAE-based interpretability produces compelling demonstrations in selected cases, current methods often fall short of the systematic reliability required for safety-critical applications. 
This suggests a necessary shift in focus from prioritizing interpretability of internal representations toward reliable prediction and control of model output.
Our work contributes to a more nuanced understanding of what mechanistic interpretability has achieved and highlights fundamental challenges for AI safety that remain unresolved.}




\maketitle

\section{Introduction}\label{sec:intro}

\textit{Mechanistic interpretability} (MI) is a form of explainable AI. It refers to a broad set of recent approaches to understanding, and ultimately controlling, the behavior of Large Language Models (LLMs). 
The core idea is to construct semantic representations -- so-called \enquote{features} -- from the neural activation patterns of the network, that are both human-interpretable and controllable. That is, the extracted features are supposed to map straightforwardly onto human concepts and should be manipulable to influence the LLM output in a meaningful way. 

The proposal is intriguing: If it is possible to faithfully map the inner workings of an LLM with its highly distributed representations into a representation that lends itself to human interpretation, \emph{and} if it is in addition possible to manipulate these interpretable features such that their role in the output is correspondingly emphasized or de-emphasized, then we have achieved a \emph{causal understanding} of the inner workings of an LLM. A causal understanding of a system is considered the ultimate goal of understanding in many scientific domains. In the case of an LLM, such a level of understanding would in addition be a boon for AI safety, since it would allow control over the LLM in a form that is familiar to the human: emphasize or de-emphasize, or even disallow, generation of text on certain topics by regulating the features that are semantically involved in the topic of interest. 

This approach has been gaining traction over the last few years, as can be seen from the contributions of different research groups \citep[e.g.,][]{Gao.2024, Gao.2025, Geiger.2025}, the growing number of review articles \citep[e.g.,][]{Rauker.2023, bereska2024mechanistic, Sharkey.2025}, and dedicated workshops on mechanistic interpretability at well-known conferences \citep[e.g.,][]{ICML.2024, NeurIPS.2025}. In the context of our investigation, however, we will focus on recent work by Anthropic. Here we can draw on a sequence of publications \citep[]{elhage2022superposition, bricken2023monosemanticity, templeton2024scaling, Durmus.2024} and an excellent open-source resource that Neuronpedia has provided for users to analyze various open source LLMs with pre-trained resources \citep[]{Neuronpedia_2023}. In many ways, Anthropic's work can be seen as representative of the applied MI research.

In \textit{Scaling Monosemanticity} \citep[]{templeton2024scaling} the Anthropic team extracts from the real-world Claude 3.0 Sonnet model a large number of features that they claim to be \enquote{generally interpretable}, and show how these features can be emphasized or de-emphasized (\textit{feature steering}) to influence their role in the LLM output. These results build on their prior work in simpler models showing how a network may internally represent input features in a \emph{superposition}, i.e.\ where a model represents more features than it has dimensions \citep[]{elhage2022superposition}. These superpositioned features can be disentangled again using so-called \textit{sparse autoencoders} (SAEs) \citep[]{bricken2023monosemanticity}. SAEs project the neural activations into a higher dimensional space where there are enough neurons to achieve a one-to-one mapping between feature and neuron. In that sense then the features are monosemantic: there is a neuron for each feature in the SAE, one can control that feature through control of the corresponding neuron, and -- ideally -- the feature is human-interpretable. 
In a recent blogpost the team has then extended this approach to identify monosemantic features across different layers of the network using so-called \textit{crosscoders} \citep[]{Crosscoder.2024}, but we will focus on the work in \citet[]{templeton2024scaling}.

In the following, we provide an initial stress-test of Anthropic's claims. We attempt to reproduce their results and test them in a similar setting under slightly different circumstances, with the overall aim to address the question of whether their approach of semantic feature extraction indeed offers a promising route for AI safety. In Section~\ref{sec:background} we first summarize the motivation underlying the approach of mechanistic interpretability. In Section~\ref{sec:AnthropicClaims} we describe Anthropic's central claims about their results regarding mechanistic interpretability. Section~\ref{sec:replication} then describes our attempts at reproducing Anthropic's findings, while Section~\ref{sec:investigation} covers our own analysis of the interpretability and steerability of some of the features we found. We discuss our results and their implications in Section~\ref{sec:Analysis}, and conclude in Section~\ref{sec:Conclusion}.

\section{Background: Mechanistic Interpretability}\label{sec:background}

Mechanistic interpretability is one approach to providing an understanding of the behavior of AI systems (mostly LLMs) by aiming 
\begin{quote}
    [...] to completely specify a neural network's computation, potentially in a format as explicit as pseudocode (also called reverse engineering), striving for a granular and precise understanding of model behavior. \citep[1]{bereska2024mechanistic} 
\end{quote}    
While there is no consensus on the precise demarcation of MI from other approaches to model interpretation (see \citet[8]{Rai.2025}; \citet[]{Saphra.2024}), MI can be broadly seen as aiming to uncover a mapping between the activations of neurons in the network and features that permit an interpretable description of what the network is doing: 
\begin{quote}
    Mechanistic interpretability seeks to understand neural networks by breaking them into components that are more easily understood than the whole. By understanding the function of each component, and how they interact, we hope to be able to reason about the behavior of the entire network. \citep[]{bricken2023monosemanticity}
\end{quote}
The approach was originally motivated by the so-called \emph{linear representation hypothesis}. The idea is that a neural network represents semantically meaningful input features as directions in its activation space \citep[]{elhage2022superposition} and that the task of understanding the network's inner operations is a matter of decoding these features again at different points within the network. Famously, \citet{Mikolov.2013} describe how a recurrent neural network language model implicitly learns certain vector-space word representations $v(\mathbf{x})$, allowing for basic algebraic operations of the activation vectors, such as $v(\text{``king"})-v(\text{``man"})+v(\text{``woman"}) \approx v(\text{``queen"})$. Ultimately, linearity may just be seen as a useful simplification to illustrate the hypothesis, holding empirically in many, but not all cases \citep[8-9]{bereska2024mechanistic}. Others, such as \citet[31]{Geiger.2025} propose to relax this restriction to include more general decompositions. In addition, it is now widely recognized that there may well be more encoded features than there are model dimensions, such that features are said to be represented in a \enquote{superposition} \citep{Arora.2018,Olah.2020,elhage2022superposition,Cunningham.2023}.

There are different approaches to discover, extract, and study functionally useful features in LLMs. One particular technique used in MI research in general, and by Anthropic in particular, are sparse autoencoders \citep[15-16]{Rai.2025}. As \citet{bricken2023monosemanticity} state, 
\begin{quote}
    [...] we use a weak dictionary learning algorithm called a sparse autoencoder to generate learned features from a trained model that offer a more monosemantic unit of analysis than the model's neurons themselves.
\end{quote}

Traditionally, an autoencoder is a network that tries to learn an approximation to the identity function through a (normally) narrow bottleneck: first encoding the input in a lower dimensional space and subsequently reconstructing the information. The main idea is that the network identifies and uses any underlying structure in the input data in order to compress the information in a lower dimensional space before reconstructing the input. In contrast to this traditional approach to autoencoders with their lower-dimensional bottleneck, the intermediate layer of a \textit{sparse} autoencoder has many more neurons than the hidden layers in the base LLM. The sparsity is enforced via added constraints -- in the case of Anthropic this is the inclusion of an $\mathit{l}_1$ penalty to the mean squared error loss \citep[]{bricken2023monosemanticity}.\footnote{For a discussion of the tendency of the $\mathit{l}_1$ norm to prefer sparse solutions in various problems, see \citep[ch. 1]{Elad.2010}.} The working assumption of MI is that within the LLM, features are represented and processed in an entangled representation, and the goal for the SAE is to \enquote{disentangle} these mixed activations by projecting them to a higher-dimensional space:
\begin{quote}
    Once the SAE is trained, it provides us with an approximate decomposition of the model's activations into a linear combination of \enquote{feature directions} (SAE decoder weights) with coefficients equal to the feature activations. \citep[]{templeton2024scaling}
\end{quote} 
While the approach of extracting features using SAEs is in principle scalable, there are practical constraints (e.g., expended resources), pragmatic decisions (e.g., particular choices of the relevant hyperparameters), and theoretical limitations that influence the quality and \enquote{completeness} of the features. For example, the largest SAE used by Anthropic in \citet{templeton2024scaling} has a dimension of around 34 million, focusing only on the residual stream activations in the middle of their model due to computational costs of the SAEs (and to prevent interference from other layers). By their own account, they believe this to be orders of magnitude short of finding a \enquote{complete} set of features, though one should note that no precise account of what would constitute a \enquote{complete set of features} is given. 

For the remainder of this paper we will adopt the technical meaning of \textit{feature} as referring to a unit in a high-dimensional layer in the SAE that is used to disentangle the model activations \citep[]{templeton2024scaling}.
Consequently, one of the important questions to grasp the notion of \textit{human-interpretability}, will be to determine the relation between these features and human concepts. \citet[]{Freiesleben.2026} offers a definition of what it means for a prediction model, such as an LLM, to represent a learned (human) concept $C$ in model part $U$:
\begin{enumerate}
    \item The model part $U$ coactivates when exhibited to instances of the concept $C$ in the input.
    \item Manipulating the model part $U$ changes the functional role of the concept $C$ in the prediction. 
\end{enumerate}
\citet[277]{Freiesleben.2026} offers the following simple example: 
\begin{quote}
    An ANN represents the concept of addition in a model part $U$ when each time it performs an addition, the part is activated and removing the unit results in ANNs no longer being able to perform common addition problems that they were previously capable of solving.
\end{quote}
If we take these criteria as a starting point, then to claim that an SAE-constructed feature is human-interpretable means that it has to reliably activate in response to a human concept in the input and its manipulation must result in changes to the corresponding concept in the output. We will adopt this view as a working hypothesis.

\section{Anthropic's Claims}
\label{sec:AnthropicClaims}
Our analysis of SAE-based features will focus on two claims by the Anthropic team that are explicitly laid out in the proof-of-concept work in \citet[]{bricken2023monosemanticity} and the follow-up work by \citet[]{templeton2024scaling}, who scale the approach up to production-level LLMs. 
Both claims form the conceptual basis for much of Anthropic's interpretability research: (1) SAEs produce interpretable features, (2) these can be used to steer the models. The research teams at Anthropic essentially propose that one is able to understand the computation of an LLM works via claim (1) and that one is then able to control the output of said computation via claim (2).

Without internal access to Claude 3.0 Sonnet (the model used in \cite{templeton2024scaling}) our analysis is necessarily based on an assumption that is widespread in the MI community, and that can also be found in Anthropic's publications \citep[see, e.g.,][]{bricken2023monosemanticity}, namely, \textit{universality}. In broad terms, universality refers to the idea that the same feature(s) can be found across different models. \citet[35]{bereska2024mechanistic} distinguish between a \textit{strong} and a \textit{weak} notion of universality. While a strong version claims that there exist some underlying, fundamental computational primitives that all models necessarily converge towards, a weak version instead suggests that the features that emerge depend on the given task, dataset, and model architecture, and different models will converge towards an optimal solution given these constraints, yielding similar features.
We adopt a weak form of universality in our analysis, as we use a publicly available model (Llama3.1-8B) and assume that our results are indicative, if not representative, of the work done by Anthropic, using their own, proprietary models. 

In the following, we lay out the two claims and show where our subsequent analysis of their validity and scope is going to interject.

\bigskip\noindent\textbf{(C1) Sparse autoencoders produce interpretable features.}\smallskip\\
We take this to be one of the most important, and most basic, claims underlying Anthropic's interpretability research program. It is made in \citet[]{bricken2023monosemanticity} -- \enquote{Sparse autoencoders produce interpretable features that are effectively invisible in the neuron basis.} -- and repeated in \citet[]{templeton2024scaling} -- \enquote{Sparse autoencoders produce interpretable features for large models}.

SAEs extract, depending on their size, millions of features per layer and researches acknowledge that there exists a wide variety of them -- varying in scope, specificity and abstraction. For example, \citet[]{templeton2024scaling} show that there exists a \enquote{Golden Gate Bridge} feature as well as a feature for \enquote{transit infrastructure}; for code related content they show features like \enquote{code error} and \enquote{addition}; and then there are very abstact features like \enquote{sycophantic praise} and \enquote{gender bias awareness}.
Given the large number of extracted features, it is in practice not possible to interpret all of them by hand. Instead, Anthropic uses an automated interpretability pipeline \citep[]{bricken2023monosemanticity} that generates descriptions of all features using a second LLM. Cunningham and colleagues summarize the basic process as follows:
\begin{quote}
    [...] the autointerpretability procedure takes samples of text where the dictionary feature activates, asks a language model to write a human-readable interpretation of the dictionary feature, and then prompts the language model to use this description to predict the dictionary feature's activation on other samples of text. The correlation between the model's predicted activations and the actual activations is that feature's interpretability score. \citep[3]{Cunningham.2023}
\end{quote}
As such, interpretations of features found by SAEs are based entirely on their activations on large text samples, and are \textit{a priori} independent of their influence on model behavior.\footnote{\citet[]{bricken2023monosemanticity} provide additional information regarding the specific implementation of their automated interpretability method in the Appendix to their work. For more information on automated interpretability, the interested reader is referred to \citet[]{Bills.2023}.}

Anthropic maintains that this approach is a strong contender for obtaining feature interpretations that track human concepts. For example, consider the following passage: 
\begin{quote}
    Many features are multilingual (responding to the same concept across languages) and multimodal (responding to the same concept in both text and images), as well as encompassing both abstract and concrete instantiations of the same idea (such as code with security vulnerabilities, and abstract discussion of security vulnerabilities). \citep{templeton2024scaling}
\end{quote}
But such enthusiasm has been met with some caution in the literature. \citet[4]{bereska2024mechanistic} note that there is no principled reason for features found in LLMs by SAEs to be humanly interpretable, and that extracted features might even become less interpretable the bigger and more complex the models become.\footnote{Note that this claim contradicts the \emph{Natural Abstraction Hypothesis}, which states that LLMs (and cognitive systems in general) converge on similar conceptual representations as humans. \citep[]{Chan.2023}.}  \citet[]{Durmus.2024} suggest that the automated interpretations should only be seen as hypotheses about the concepts the features might correspond to. In other writing, Anthropic's own assessment is also more measured:
\begin{quote}
    Another issue is that we don't believe our features are completely monosemantic (some polysemanticity may be hiding in low activations), nor are all of them necessarily cleanly interpretable. \citep[]{bricken2023monosemanticity}
\end{quote}

This points to the crux of understanding Anthropic's claims: On the one hand they acknowledge the sharp distinction between their extracted features and human interpretable concepts, and on the other hand it appears to be the driving idea of their mechanistic interpretation that the features align with human-interpretable concepts in order to gain insights into the inner workings of LLMs.
There is, of course, no inconsistency in claiming that many features match human concepts and some features are not interpretable. But then we need to understand how we identify the features that indeed track concepts, how one can manipulate them, and how they interact with the non-interpretable features.
One can even take a closer look at the interpretable features themselves. As features can be very complex -- \enquote{encompassing both abstract and concrete instantiations of the same idea} \citep{templeton2024scaling} -- some overlap between them is inevitable. However, as the number of features is fixed by the SAE and no restrictions are imposed on the allowed similarity between them, this can lead to individually interpretable features with no clear distinction between them.
Moreover, recall that automated interpretations are used to generate the human-interpretable description of features from the lists of activation patterns on text samples. Could features, therefore, correspond to activation patterns that depict abnormal behavior in and of themselves, without showing up in the automatically generated interpretations? Could these automatic interpretations also mask the context-dependence of activation patterns, when at the same time the performance of transformer-based LLMs is so heavily linked to their ability to incorporate contextual information?

\bigskip\noindent\textbf{(C2) Features can be used to steer models.}\smallskip\\
This second claim is made in \citet[]{bricken2023monosemanticity} -- \enquote{Sparse autoencoder features can be used to intervene on and steer transformer generation.} -- and repeated in \citet[]{templeton2024scaling} for large models: \enquote{Features can be used to steer large models [...]. This extends prior work on steering models using other methods [...].}

\citet[]{templeton2024scaling} show how this works for Claude 3.0 Sonnet. The outputs of the SAE are spliced back into the transformer. With each neuron of the SAE representing one feature, it is possible to increase the output for each feature by some multiple of the original value (a step called clamping). Clamping the feature \enquote{transit infrastructure} to $5\times$ its maximum activation value changes the transformers answer to the question \enquote{What's the best way to get to the grocery store down the street? Be brief.} from \enquote{1. Walk.} to \enquote{1. Walk across the bridge} \citep[]{templeton2024scaling}.  

The claim that the features produced by SAEs can be used to successfully influence the behavior of practically employed LLMs plays an important double role in Anthropic's work. On the one hand, it is crucial for the overarching motivation behind their work on MI, because steering has direct implications for model safety and ethical considerations. For example, \citet[]{Durmus.2024} show that feature steering can reduce (or increase) social biases: clamping the feature \enquote{Neutrality and impartiality} by a positive value reduces the bias score on the \textit{Bias Benchmark for Question Answering}.
  
On the other hand, manipulability of the features with respect to the output of the model is taken as a significant indicator of representational correctness: \enquote{To verify our feature interpretations, we ran qualitative feature steering experiments, where we artificially dialed up and down various features to see if they changed model outputs in intuitive ways} \citep[]{Durmus.2024}.
As such, the fact that features can be employed to influence model behavior is used as practical confirmation of the underlying methodology and as a goal to enhance AI safety.

Together, both perspectives place a significant burden on the steering capabilities of features. The large number of features makes an automatic interpretation process necessary, leading us to rely heavily on steering as a fail-safe as well as an important tool for AI safety. This raises questions about the interplay between activation, interpretation and steering. 
Automated interpretation could result in a feature description not corresponding well with its activation. How does this translate into steering behavior? Even if they correlate, could there be cases where interpretation and steering could lead to a false conception of a feature?
Different steering coefficients can be chosen to adjust the influence of each clamped feature on the output. Given that we can specify the exact steering coefficients for each feature, is it possible to control the magnitude of the output change with such precision? Are the magnitudes similar for different features, or does each feature have its own sweet spot? Could the position of the layer on which the SAE is trained also influence this sweet spot behavior?
Finally, while it might be helpful to carefully regulate the influence of some features, could there also be unexpected side effects of feature steering? That is, for some steering coefficient, the output may not change in the way the activation and description led us to believe.

\section{Replicating Anthropic's Results}\label{sec:replication}
Our goal is to critically examine the two claims underlying Anthropic's recent work on interpretable features in LLMs. As a starting point, we will first reproduce the relevant basic results found in \textit{Scaling Monosemanticity} \citep{templeton2024scaling} on a publicly available model.

In order to reproduce Anthropic's results -- and later test their generalizability and expand on Anthropic's findings -- we use an open source LLM and SAE.
Specifically, we use Llama 3.1 with 8 Billion parameters as our base LLM -- a lightweight and fast version of Llama. Thanks to Neuronpedia\footnote{Neuronpedia can be accessed through its website (\url{https://www.neuronpedia.org/}). Most of its functions -- especially the ones relevant in this paper -- are also accessible through an API.} we have access to various pre-trained SAEs for Llama \citep[]{Neuronpedia_2023}. All of the following results are based on all 32 SAEs (one for each layer of Llama 3.1) called \textit{Llama Scope} with an 8-fold multiplication, i.e. 32k features per layer \citep{He_2024}. To make our results comparable to Anthropic's approach, we selected SAEs that were trained on the residual stream between layers rather than on the multilayer perceptrons (MLPs) in each layer. This enables the SAEs to access the full information flow between layers, rather than just the output of a single MLP at a specific point. We will refer to the SAEs by the layers behind which they were trained. For example, \enquote{SAE in layer 18} refers to the SAE that was trained on the residual stream after the output of the MLP in layer 18 was added back in.  Additionally, we ensure reproducibility by fixing all of the parameters above, as well as seed, temperature, etc. throughout the paper.
The automated interpretation of features was done by o4-mini \citep{hijohnnylin_2025} based on an implementation by \citet{OpenAI_2025b}. For a detailed list of our parameters, see Appendix \ref{app:SetupDetails}.

With this setup, we can extract feature activations for any input and steer the LLM's output using arbitrary steering coefficients for all features of all layers.  Below, we present one result from each of our tests and provide additional data in the Appendix. However, unlike Anthropic, we do not have the resources to be exhaustive or even properly representative. Instead, we present exemplary findings to caution against hasty generalizations of positive examples often cited in the literature.
In the following, we show that we are able to replicate Anthropic's results qualitatively. We give an example similar to the ones \citet[]{templeton2024scaling} use, check its specificity, and its steering capabilities. Finally, we comment on the general trends we can observe when working with features.

\subsection{Replicating Interpretable Feature Extraction and Activation}
As a first step, we show that there exist interpretable features that activate in the appropriate contexts. To do so, we need to find a feature that is similar to the \enquote{Golden Gate Bridge} feature in \citet{templeton2024scaling}: the feature should have a clear-cut meaning (as opposed to a feeling or qualitative judgment) and a \enquote{normal} activation pattern. In our case, the SAEs have over $30,000$ features for each layer. As we describe in Section \ref{subsec:steering}, the most stable features with regard to activation and steering can be found in the middle layers of the network. To illustrate and reproduce Anthropic's results, we select feature $9463$ in layer $18$ as an example. It meets the above criteria and is auto-interpreted as \enquote{mentions of coffee and related terms}.
In the following, we will abbreviate feature number $Y$ in layer $X$ as ($X/Y$); so, the feature \enquote{mentions of coffee and related terms} has the unique identifier ($18/9463$).

As we depict in table \ref{tab:activation_english}, given a longer text passage on coffee, this feature activates on words like \enquote{coffee} and \enquote{espresso}, as well as words related to this context. The closer the words are to \enquote{coffee} itself, the stronger the activation, though it does not pick up on every coffee-related term (e.g., \enquote{portafilter}, \enquote{ristretto}, \enquote{crema}). In comparison to others, this feature activation is fairly specific: In Section \ref{subsec:ContextDependence} we compare three features (including this one) with varying context sensitivity and this one is not particularly context sensitive.

\begin{table}[ht!]
    \small
    \captionsetup{font=small}
    \centering
    \begin{tabular}{p{.96\textwidth}} \hline
         As the bar\redhighlight{red!18.44}{\strut ista} carefully grinds the aromatic \redhighlight{red!49.01}{\strut espresso} beans and tamps them into the portafilter before pulling a silky shot of ristretto, the bustling \redhighlight{red!26.11}{\strut café} fills with the comforting scent of roasted Arab\redhighlight{red!23.02}{\strut ica}, while patrons sip their c\redhighlight{red!23.02}{\strut app}\redhighlight{red!10.89}{\strut ucc}\redhighlight{red!25.99}{\strut inos} from ceramic mugs, chatting about fair-trade plantations, lat\redhighlight{red!34.65}{\strut te} art, crema, and the timeless ritual that makes \redhighlight{red!84.65}{\strut coffee} not just a beverage but a global symbol of morning routines, social connection, and mindful indulgence.\\ \hline
    \end{tabular}
    \caption{Activation of feature \enquote{mentions of coffee and related terms} (18/9463) on English text. The feature related to coffee activates on and around coffee terms. The closer the words are to coffee the higher the activation. Higher opacity indicates stronger activation.}\label{tab:activation_english}
\end{table}

Similar to the examples provided by Anthropic, this feature activates in the context of coffee in multiple different languages, such as German (table \ref{tab:activation_german}) and Japanese (table \ref{tab:activation_japanese}).\footnote{The text examples are automatic translations of our English text. The translations were generated by an ML model that is distinct from the Llama 3.1 model that was evaluated in this study, and checked by native speakers.} This shows that it is a multilingual feature, which abstracts to some degree from language specific words and activates in a narrow margin on and around the topic of coffee. This is an important characteristic if one intends to steer the feature in a model independently of language.

\begin{table}[ht!]
    \small
    \captionsetup{font=small}
    \centering
    \begin{tabular}{p{.96\textwidth}} \hline
        \textbf{German:}
        W\"ahrend der Bar\redhighlight{red!16.83}{\strut ista} die aromatischen Es\redhighlight{red!28.09}{\strut press}\redhighlight{red!16.09}{\strut ob}ohnen sorgf\"altig mahlt und sie in den Siebtr\"ager dr\"uckt, bevor er einen seidigen Schuss Ristretto zieht, erf\"ullt sich das gesch\"aftige \redhighlight{red!35.64}{\strut Caf\'e} mit dem wohligen Duft von ger\"ostetem Arab\redhighlight{red!16.96}{\strut ica}, w\"ahrend die G\"aste ihren C\redhighlight{red!22.40}{\strut app}\redhighlight{red!8.6}{\strut ucc}ino aus Keramiktassen schl\"urfen und sich \"uber Fair-Trade-Plantagen, Lat\redhighlight{red!17.95}{\strut te} Art, Crema, und das zeitlose Ritual unterhalten, das K\redhighlight{red!53.71}{\strut aff}\redhighlight{red!50.74}{\strut ee} nicht nur zu einem Getr\"ank, sondern zu einem weltweiten Symbol f\"ur Morgenroutine, soziale Kontakte und bewussten Genuss macht.\\ \hline
    \end{tabular}
    \caption{Activation of feature \enquote{mentions of coffee and related terms} (18/9463) on German text. The feature related to coffee activates on and around coffee terms (even in German). The closer the words are to coffee the higher the activation.}
    \label{tab:activation_german}
\end{table}

\begin{table}[ht!]
    \small
    \captionsetup{font=small}
    \centering
    \begin{tabular}{p{.96\textwidth}} \hline
        \textbf{Japanese:}
        \begin{CJK}{UTF8}{min}
        バリスタが香り高いエスプレッ\redhighlight{red!35.89}{\strut ソ}豆を丁寧に挽き、ポルタフィルターに詰めてから、なめらかなリストレットを注ぎ 、やかなカ\redhighlight{red!49.26}{\strut フェ}には心地よい香りが広がります。その間、常連客は陶器のマグカップでカプチー\redhighlight{red!24.38}{\strut ノ}を飲みながら、フェアトレード、ラテアート、クレマ、そしてコー\redhighlight{red!75.74}{\strut ヒ}\redhighlight{red!52.23}{\strut ー}\redhighlight{red!4.49}{\strut を}巡る儀式について会話します
        \end{CJK}\\ \hline
    \end{tabular}
    \caption{Activation of feature \enquote{mentions of coffee and related terms} (18/9463) on Japanese text. The feature related to coffee activates on and around coffee terms (even for Japanese characters). The closer the characters are to coffee the higher the activation.}
    \label{tab:activation_japanese}
\end{table}

\subsection{Replicating Feature Activation Specificity}
The next step is to check the specificity of our coffee-feature. We want our feature to activate only in the context that its description mentions. The further away the content of a text is to the feature, the lower its activation should be. While we are not able to test the feature's activation on a large corpus of data, we can reproduce the approach by Anthropic on a small scale. \citet[]{templeton2024scaling} propose four text categories with regard to feature activation: (0) The feature is completely irrelevant throughout the context; (1) The feature is related to the context, but not near the highlighted text or only vaguely related; (2) The feature is only loosely related to the highlighted text or related to the context near the highlighted text; (3) The feature cleanly identifies the activating text.

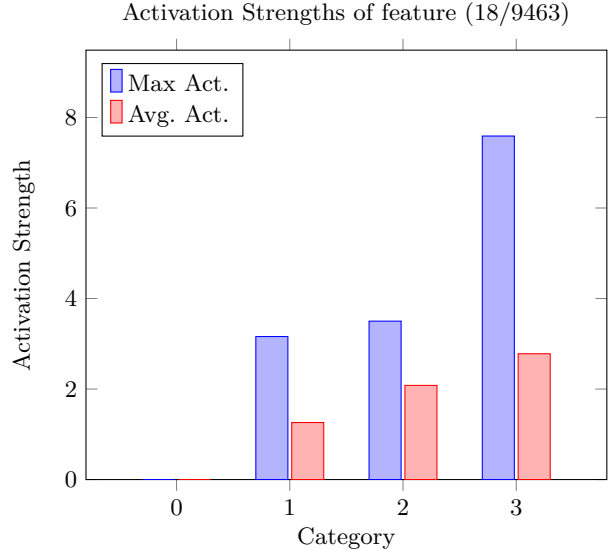
\begin{figure}[ht!]
    \small
    \captionsetup{font=small}
    \centering
    \begin{tikzpicture}
        \begin{axis} [ybar = .05cm,
            bar width = 12pt,
            title={Activation Strengths of feature (18/9463)},
            xtick = data,
            ylabel= Activation Strength,
            ylabel style={at={(0.06,0.5)}},
            xlabel= Category,
            enlarge y limits = {value = .25, upper},
            enlarge x limits = {abs = .8},
        	legend pos=north west,
            legend image code/.code={
                \draw [#1] (0cm,-0.1cm) rectangle (0.15cm,0.2cm);},
            ]
            \addplot coordinates {(0,0) (1,3.16) (2,3.5) (3,7.59)};
            \addplot coordinates {(0,0) (1,1.26) (2,2.08) (3,2.78)};
         
            \legend {Max Act., Avg. Act.};
        \end{axis}
    \end{tikzpicture}
    \caption{Activation of feature \enquote{mentions of coffee and related terms} (18/9463) on sentences belonging to the four specificity categories (category 0 is not related and category 3 is very related; see Appendix \ref{app:AnthropicCategoriesSentences}). The more related the context in a category is to \enquote{coffee}, the higher the activation.}\label{tab:Activation OnCategories}
\end{figure}

We select multiple sentences that should fall into each of these categories based on our feature's description. Here is an example for each of the levels (see Appendix \ref{app:AnthropicCategoriesSentences} for a detailed list): (0) \enquote{The quantum fluctuations in vacuum energy remain a mystery in theoretical physics}; (1) \enquote{Many people gather in social spaces like bistros and lounges to chat before work}; (2) \enquote{Sometimes she opted for a macchiato instead of her usual hot chocolate}; (3) \enquote{He sipped a creamy cappuccino topped with frothed milk at the busy coffeehouse.} We can then extract the maximum activation on these sentences as well as the average over all non-zero activations (\enquote{non-zero} to compensate for different numbers of tokens). We see in Figure~\ref{tab:Activation OnCategories} that both the maximum as well as the average activations grow from one categroy to the next, showing that our coffee-feature activates with specificity in the context of \enquote{mentions of coffee and related terms}.

\subsection{Replicating Feature Steering in Output}\label{subsec:steering}
Finally, we need to check for the influence this feature can exert on the output of the LLM. In order to achieve an effect, a so-called steering coefficient can be used to clamp a specific feature. In that case, the output of the SAE for the clamped feature consists of the baseline output of the SAE plus that output multiplied by the steering coefficient. This method ensures that a steering coefficient of $0$ does not change the model's behavior.
As shown in table \ref{tab:clamping_2x_coffee}, we clamp the coffee-feature with a steering coefficient of 2 and use the input \enquote{My favorite drink is}. While the baseline model outputs alcoholic beverages, the steered model instead generates coffee-related outputs. In the following, we always highlight in bold the interesting passages that change for the steered model.
\begin{table}[ht!]
    \small
    \captionsetup{font=small}
    \centering
    \begin{tabular}{|p{.12\textwidth}|p{.82\textwidth}|}\hline
         Steering & My favorite drink is... \\ \hline\hline
         Baseline & ... a \textbf{margarita} but my favorite cocktail is a mojito. I have a lot of different \textbf{recipes for mojitos}, but this one is my favorite. It's simple and packs a punch. The name of this drink, \enquote{\textbf{Mojito},} means \enquote{little drink} in Spanish. It's made with [...] \\
         \hline
         Coeff $2$ & ... \textbf{coffee}. I usually \textbf{drink my coffee black}, but sometimes I like to add a little milk or cream to it. I also like to add sugar or sweetener to my \textbf{coffee}.
        \textbf{Coffee} is a brewed beverage with a distinct aroma and flavor. It is one of the most popular drinks in the world, and it has been [...] \\
        \hline
    \end{tabular}
    \caption{Steering capability of feature \enquote{mentions of coffee and related terms} (18/9463). Positively clamping this feature changes the output from alcoholic beverages to coffee-related content.}
    \label{tab:clamping_2x_coffee}
\end{table}
        
As we show in table \ref{tab:clamping_-2x_coffee}, this behavior is consistent for positive and negative clamping. If we clamp the feature negatively (we use a steering coefficient of $-2$), we observe the opposite behavior to earlier: the suppression of coffee-related output. The change to \enquote{100\% natural} might not be entirely grammatically correct, although the supplements mentioned in the second sentence are often marketed as \enquote{natural}.
\begin{table}[ht!]
    \small
    \captionsetup{font=small}
    \centering
    \begin{tabular}{|p{.12\textwidth}|p{.82\textwidth}|}\hline
        Steering & An important part of a Swedish breakfast is grown and harvested mainly in Brazil and is called... \\ \hline\hline
        Baseline & ... 100\% \textbf{Arabica coffee}. It comes from the \textbf{coffee bean} which is processed, \textbf{roasted and ground} to make the drink we all know and love. The \textbf{coffee plant} is a tropical evergreen shrub that grows up to 5 meters high if left unpruned [...] \\
        Coeff $-2$ & ... 100\% \textbf{natural}. This includes the essential vitamins and minerals that your body needs. The protein content of this product is very high, which makes it a good choice for people who are looking to gain muscle mass or lose weight. It also has a low glycemic index, which means that it will not cause spikes in your [...] \\ \hline
    \end{tabular}
    \caption{Steering capability of feature \enquote{mentions of coffee and related terms} (18/9463). Negatively clamping this feature changes the output from coffee to some generic product with health benefits.}
    \label{tab:clamping_-2x_coffee}
\end{table}

Note that Anthropic focuses their work on the \enquote{middle layer} of the model because they expect it to contain \enquote{interesting, abstract features} \citep[]{templeton2024scaling}. In our analysis, we also considered features and their steering capabilities in earlier and later layers of the network. While we do not have the resources for a thorough quantitative assessment, we did also qualitatively find that feature steering works best for layers in the middle of the network, whereas steering in early layers tends to result in chaotic model behavior, and steering in later layers tends to have little influence on model output. 

Regarding the size of useful steering coefficients, we observed a qualitatively similar \enquote{feature steering sweet spot} as described by Anthropic.\footnote{\citet[]{Durmus.2024} analyzed $29$ different features and tested the capabilities of the steered model for different steering coefficients using the MMLU (Massive Multitask Language Understanding) benchmark. They observe that a steering coefficient in the range between $-5$ and $5$ does not significantly impact model capabilities. Within the \enquote{sweet spot} between $-5$  and $5$ the steered model scored similarly well on this benchmark, with a steering coefficient of $0$ yielding the best accuracy.} However, their results only show if a steering coefficient negatively impacts the performance of a model (e.g., the output only consists of gibberish), not if a steered model shows any change in line with the feature used for steering. In other words, their test does not allow any statement about how well the features actually work -- maybe they score so well because they have little impact on model behavior in the first place. 

We found that in general, the steering quality depends on the specific feature and layer, and on the size of the steering coefficient. Features in the middle of the network provide the best compromise between impact and specific control that we found. \\

Overall, we were able to replicate Anthropic's base results on interpretable features in LLMs using open-source tools (Llama 3.1 and publicly available SAEs). In short, the basic mechanics of feature extraction, activation, and steering work as claimed by \citet{templeton2024scaling}, thus setting the stage for our deeper investigation of whether these findings generalize reliably.

\section{Do the Findings Generalize?} \label{sec:investigation}

In Section~\ref{sec:replication} we showed that we are able to reproduce Anthropic's basic results. We now turn to the stronger promise associated with this work that suggests that such extraction of interpretable features can provide model understanding and, through feature steerability, a tool for AI safety. This requires the proof of concept to generalize across a variety of features, contexts and steering settings. We start with a closer look at the types of features that are extracted (Section~\ref{subsec:scope}), at the similarity and distinctness of the extracted features (Section~\ref{subsec:Similarity}), their context-dependence (Section~\ref{subsec:ContextDependence}) and at what one might call \enquote{spurious activations} (Section~\ref{subsec:CuriousActivation}). 
Finally, Section~\ref{subsec:Coherence} focuses in more detail on the relationship between steering and output results. This serves as a check of the specific behavior observed in Section~\ref{subsec:steering}.

In the following analysis we rely on (generally, multiple) examples to illustrate our points, working on the assumption that if the proposed approach to MI is supposed to be a viable candidate for safety-critical settings, then even a few easily detectable shortcomings are cause for concern.  

\subsection{Do the Features Match Their Interpretations?}\label{subsec:scope}
We first investigated how representative features such as our \enquote{mentions of coffee and related terms} are, or whether this is a particularly well-behaved example. 

As we show in the following, there are many other easily interpretable features -- some of them are considerably more abstract than our original example -- that show consistent activation patterns and model-steering capabilities. However, we also encountered many features that formally resemble those that can be used for steering, but for which we have no meaningful interpretation. For these features, neither their activation on small text samples, nor their steering capabilities match their alleged interpretation.\footnote{Note that these are different from \enquote{dead} features that never activate or that have no effect when steered.} 

\subsubsection{Example Features with Successful Interpretations}\label{subsubsec:GoodFeatures}
\noindent\textbf{(1) \enquote{expressions of regret, mourning, and sadness} (17/16706)}\smallskip\\

As expected, this feature activates strongly on negative emotions, such as \enquote{regret}, \enquote{grief}, \enquote{sorrow}, and \enquote{sadness}. Furthermore, we see lower-level activations for related expressions like \enquote{heart}, \enquote{mourning}, and \enquote{ache}/\enquote{aching}. The activation pattern aligns well with the interpretation of the feature, which is considerably more abstract than the coffee-example above. 

When clamping this feature at a coefficient of $10$, we see consistent steering of the model output towards negative emotional states, no matter the tone of the original input. In one example (Input: \enquote{Summer is the time of the year where I}), the model output shifts from a happy and grateful reply towards a sorrowful and averse one. Similar behavior can be observed with the input \enquote{The best thing that ever happened to me was}: While the prompt should strongly influence the model towards a positive output, the steered model generates a sad and melancholic story. The exact activation and steering responses can be found in Appendix \ref{app:GoodFeatures_regret}.

\bigskip\noindent\textbf{(2) \enquote{instances of the phrase \enquote{would have} or similar expressions indicating hypothetical or conditional situations} (19/12491)}\smallskip\\
While the interpretation of this feature is rather abstract, all activations in our sample text are consistent with this interpretation. The highest activations appear for \enquote{have} following \enquote{would}, whereas \enquote{would be} and \enquote{could have} show lower activations. This shows that the feature activates on specific tokens that are tightly linked to the more abstract subjunctive form of text.

A similar behavior can be observed for steering. When steering the model by a coefficient of $15$, the output consistently changes from simple, straightforward answers to more complicated outputs including a range of hypothetical/conditional statements consistently starting with \enquote{would} and \enquote{would have}. This is not only true for open-ended inputs such as \enquote{I am going to}, but also for more factual prompts, e.g., \enquote{Photosynthesis is characterized by}.

Despite the abstractness of this feature, both the activation pattern and the steering capabilities seem to align well with its interpretation. See Appendix \ref{app:GoodFeatures_WouldHave} for details.

\bigskip\noindent\textbf{(3) \enquote{references to lists and list-making processes} (16/24361)}\smallskip\\
This feature activates strongly on words such as \enquote{checklist} and \enquote{list}, and additionally on terms that provide context to these words. Furthermore, we observe activations on related descriptions such as \enquote{smaller note with labeled items}. 

When clamped with a steering coefficient of $15$, the output of the model based on the prompt \enquote{Photosynthesis is characterized by} changes from a standard description to a labeled list. Additionally, the model included the word \enquote{list} twice in its output. For the input \enquote{Let us talk about} and a steering coefficient of $5$, the model generates a modified answer which includes a list-making process. 

Again the interpretation of this feature is consistent with the observed activation pattern and steering capabilities. See Appendix \ref{app:GoodFeatures_List} for details.

\subsubsection{Examples of Mis-interpreted Features}\label{subsubsec:BadFeatures}
\noindent\textbf{(1) \enquote{references to exposure to hazards or influential experiences} (18/14711)}\smallskip\\
This feature has a rather broad interpretation, referring not only to exposure to hazards, but influential experiences in general. Accordingly, the activation pattern (see table \ref{tab:activation_exposure}) not only shows high values for \enquote{exposure in hazardous environments}, but also \enquote{exposure to new ideas} and similar contexts. While this might still be considered as an \enquote{influential experience}, we show that the same feature activates strongly for exposure in the context of photography -- a context that hardly fits the proposed feature interpretation. 

\begin{table}[ht!]     
    \small     
    \captionsetup{font=small}     
    \centering     
    \begin{tabular}{p{.96\textwidth}} \hline
        Prolonged \redhighlight{red!69.49}{\strut exposure} \redhighlight{red!21.08}{\strut to} loud noise can cause permanent hearing damage, which is why safety guidelines often stress limiting \redhighlight{red!71.19}{\strut exposure} in hazardous \redhighlight{red!4.18}{\strut environments}. At the same time, early \redhighlight{red!69.92}{\strut exposure} to new ideas and influential \redhighlight{red!18.64}{\strut experiences} can shape a person's worldview in powerful ways. Outside of health and personal growth, the concept of \redhighlight{red!100}{\strut exposure} also applies \redhighlight{red!9.8}{\strut to} art: the photographer adjusted the camera settings \redhighlight{red!8.79}{\strut to} get the perfect \redhighlight{red!61.86}{\strut exposure} for the sunset shot, ensuring that the light and colors were captured just right.\\ \hline
    \end{tabular}
    \caption{Activation of feature \enquote{references to exposure to hazards or influential experiences} (18/14711) on English text. The feature activates on the term \enquote{exposure} regardless of the context.}
    \label{tab:activation_exposure}
\end{table}

For a steering coefficient of $-10$ the feature is capable of influencing the model behavior in both contexts. In table \ref{tab:clamping_-10_exposure-a} we show that the protection from chemical experiments is less prominent than one would expect. However, in table \ref{tab:clamping_-10_exposure-b} one can see that \enquote{exposure} is also suppressed in the context of photography, which is not to be expected from the description.  

The failure of this and the following features, of course, may have resulted from many different sources: Is the feature itself in fact an agglomeration of several different human concepts related in one way or another to \enquote{exposure}, or did the automated interpretation incorrectly describe the feature, or did the steering capability fail?

\begin{table}[ht!]
    \small
    \captionsetup{font=small}
    \centering     
    \begin{tabular}{|p{.12\textwidth}|p{.82\textwidth}|}\hline
        Steering & When doing chemical experiments... \\ \hline\hline
        Baseline & ... in the lab, you should always wear \textbf{protective glasses to protect your eyes} from any splashes or spills. If you don't have access to protective glasses, you can use a pair of goggles instead. Goggles are made of plastic and have a clear lens that covers your entire eye area.\newline They are designed to \textbf{protect} your eyes from chemicals, dust, [...] \\
        Coeff $-10$ & ... in the lab, \textbf{you should always be careful} to pour out your chemicals properly. If you are making a solution, make sure you know what kinds of acids and bases you are mixing together so that you can calculate the pH of the solution. This is important because some chemicals will react with each other when they are mixed together.\newline If you are mixing two [...] \\ \hline
    \end{tabular} 
    \caption{Steering capability of feature \enquote{references to exposure to hazards or influential experiences} (18/14711). Negatively clamping this feature changes the output from explicitly mentioning \enquote{protection} to only including general descriptions of chemical procedures.} 
    \label{tab:clamping_-10_exposure-a} 
\end{table}

\begin{table}[ht!]
    \small
    \captionsetup{font=small}
    \centering     
    \begin{tabular}{|p{.12\textwidth}|p{.82\textwidth}|}\hline
        Steering & There are multiple settings that a photographer can adjust to change the... \\ \hline\hline
        Baseline & ... \textbf{exposure} of their image. These include shutter speed, aperture, and ISO. The shutter speed is the length of time that the camera's shutter is open, allowing light to enter the lens and \textbf{expose the image sensor}. A faster shutter speed will result in a brighter image, while a slower shutter speed will result in a darker image.\newline The aperture is the [...] \\
        Coeff $-10$ & ... \textbf{way a picture looks}. These include shutter speed, aperture, and ISO. The shutter speed is the length of time the shutter is open, and the aperture is the size of the lens opening. A faster shutter speed will make \textbf{a picture look blurry}, while a slower one will make it appear blurry [\textit{sic}].\newline A photographer can also adjust the ISO to change how [...] \\ \hline
    \end{tabular} 
    \caption{Steering capability of feature \enquote{references to exposure to hazards or influential experiences} (18/14711). Negatively clamping this feature suppresses the term \enquote{exposure} even though it is not contextualized with hazards or influential experiences, but photography.} 
    \label{tab:clamping_-10_exposure-b} 
\end{table}

\bigskip\noindent\textbf{(2) \enquote{references to coffee} (29/30028)}\smallskip\\
This particular feature does not only activate on \enquote{coffee} and related terms, but on the token \enquote{coff} more generally. This leads to activations in our sample text on \enquote{coffee shop} as well as \enquote{coffin}. 

As we can show by clamping this feature by a coefficient of $-20$, the steering capabilities of this feature are compatible with the observed activation pattern: Giving the input \enquote{He told me that, at the funeral, five people lowered the}, the baseline model immediately generates text that revolves around \enquote{coffin}. The steered model, however, omits any reference to \enquote{coffin}, instead using related words like \enquote{casket}. See Appendix \ref{app:BadFeatures_coffee} for details.

We observed a similar behavior for the feature \enquote{the word \enquote{Morocco} and its variations or related concepts} (18/6893). It does not only activate on the word \enquote{Morocco}, but more general on the token \enquote{mor}, as well as words containing these letters like \enquote{morsel} and \enquote{morals}. Steering also shows an influence on all tokens containing \enquote{mor} not just \enquote{Morocco}. The activation pattern and steering results for this feature can be found in Appendix \ref{app:BadFeatures_Morocco}.

This suggests that these features refer neither to the concept of coffee (or coffin), nor to Morocco, but rather to the tokens \enquote{coff} and \enquote{mor}. If that is the case, then this undermines the claim of interpretable feature extraction more generally, since the goal was to extract features akin to human concepts, not ones that represent character sequences.

\subsection{Does a Similar Interpretation Imply Similar Activations and Similar Steering Effects?}\label{subsec:Similarity}
 
When analyzing the feature \enquote{mentions of coffee and related terms}, as described in Section~\ref{sec:replication}, we encountered many other features with similar interpretations, even within one layer. If distinct features actually refer to the same concept, one would expect them to behave identically. Conversely, if two features share a similar interpretation, but their associated activation patterns and steering capabilities differ greatly, one ought to be skeptical about whether they refer to the same concept as the automated interpretation suggests.

To give some sense of the kind of activations -- and related problems -- we will talk about in the next two Sections, we systematically checked the activation  patterns of all features containing the word \enquote{coffee} on coffee-related terms (see Online Resource 1).
Three general observations can be made. First, all \enquote{coffee}-features are active for at least a single token in our test set of coffee-related terms.
Second, one can see that activations grow larger for deeper layers of the model.
Third, for multi-token words like \enquote{cappuccino} or \enquote{ristretto} most of the features do not activate on all tokens but only on a subset. This subset is relatively stable across all features. For example, all features we checked that are active for the word \enquote{cappuccino} only activate on the token \enquote{app}.
 
However, we also see some noteworthy behavior of features. Why, for example does the general sounding feature \enquote{words related to coffee} (27/1729) only activate for \enquote{frappuccino} and \enquote{arabica} and nothing else? While all other coffee drinks get at least some amount of activation, why then do only 5 features activate on \enquote{macchiato}? And finally, why does the feature \enquote{sentences that discuss coffee experiences and related cultural practices} (27/4442) activate, e.g., on \enquote{espresso}, \enquote{latte}, and \enquote{cappuccino}, and \enquote{references to coffee and its cultural significance} (27/11953) does not, while both of them activate similarly on \enquote{arabica}, \enquote{robusta}, and \enquote{roastery}?

The next two Sections will discuss in more detail whether and how we can rely on feature descriptions. To better understand the similarity of features in the following, we will first consider similar features in a single layer, and then similar features that appear in different layers.

\subsubsection{Similar Features in a Single Layer}\label{subsubsec:SimilaritySingleLayer}

Given the construction of the SAEs, it is not surprising that -- depending on the size of the SAE -- one might have a certain amount of overlap of features. As \citet[]{templeton2024scaling} describe, features in smaller SAEs tend to \enquote{split} into more fine-grained features in larger SAEs. For example, a single \enquote{San Francisco} feature found in the 1M SAE splits into two distinct features in the 4M SAE, and into eleven fine-grained features in the case of a 34M SAE. Intuitively speaking, the larger SAE is like a stronger microscope, allowing for a more detailed analysis of the feature space. This implies, however, that these fine-grained features -- while close in feature space and semantically related -- are still distinct from one another.

However, as we show in the following, this is not always the case. For example, take all coffee-related features in layer 18. As depicted in table~\ref{tab:ConfusionMatrixLayer18Coffee}, there exist five different features that broadly refer to coffee, all of which were found through a single SAE. 

If this is a case of feature splitting where a \enquote{coffee-cluster} was split up due to the size of the SAE used, then we would expect very similar -- but not identical -- coffee-related behavior across these features. This means that for a set of coffee-related words, one would expect most of the features to be activated frequently. However, there should be some phrases that only fit into the realm of a single fine-grained feature and consequentially only that feature should activate.
Alternatively, this seemingly large overlap of features could be due to the automatic and imperfect generation of interpretations and might not represent the actual differentiating power of the underlying features.

To analyze this possible discrepancy, we checked the feature activations on different coffee-related terms to see if there is a difference. Specifically, we chose ten words per feature that should (based on the feature's description) lead to a high activation of said feature. Because there are differences in the descriptions, we choose words that are specific for one category instead of being generally about coffee. For example, words like \enquote{WiFi}, \enquote{Coffeehouse playlist}, and \enquote{Urban roast bar} fit well to the description of feature (18/17305): \enquote{references to cafes and coffee culture}. While words like \enquote{Sauvignon Blanc}, \enquote{Cold brew}, and \enquote{Pinot Noir} are supposedly much better suited to generate a high activation for feature \enquote{references to drinks, particularly wine and coffee} (18/7546). For a full list of words, see the Appendix~\ref{subsubsec:CategoryTerms}. 

Of course, some overlap is inherent for these five features. For example, for the very general feature \enquote{mentions of coffee and related terms} (18/9463) a lot of activation on all of the categories is to be expected.
We recorded the activation of all features for the phrases in all categories. We then rescaled all activations of each feature between $0$ and $1$ by its maximal activation. Then the activations were rescaled across all terms in each category. Finally, we summed the activations for each feature over the terms of one category to compare the discriminatory power between the features (see table~\ref{tab:ConfusionMatrixLayer18Coffee}). For fully disjoint features we would expect activations only on the diagonal. This would mean a feature exclusively activates on the words specifically chosen for it and on nothing else. But even in the case of these five related features we would expect the highest activations to be found on the diagonal. This would mean that, while all of the features have some overlap regarding their activation on coffee-related terms, they still show some discriminatory power by activating more strongly on terms form their fine-grained region instead of on generally coffee-related terms.

    \begin{table}[ht!]
        \renewcommand{\arraystretch}{1.2}
        \small     
        \captionsetup{font=small} 
        \centering
        \begin{tabular}{>{\arraybackslash}m{.4\textwidth}|c|c|c|c|c|}
            & \multicolumn{5}{c|}{Terms for Feature} \\
            & 7546  & 9463     & 15276   & 17350  & 28590 \\ \hline
             \enquote{references to drinks, particularly \newline wine and coffee} (18/7546)       & \gcolor{11.48} 0.48  & \gcolor{0} 0          & \gcolor{26.79} 1.12     & \gcolor{3.59} 0.15       & \gcolor{7.89} 0.33    \\ \hline
            \enquote{mentions of coffee and \newline related terms} (18/9463)                      & \gcolor{45.22} 1.89 & \gcolor{48.33} 2.02     & \gcolor{34.69} 1.45     & \gcolor{59.81} 2.5     & \gcolor{99.52} 4.16 \\ \hline
            \enquote{references to coffee and coffee- \newline related experiences} (18/15276)      & \gcolor{80.86} 3.38 & \gcolor{43.3} 1.81     & \gcolor{20.57} 0.86      & \gcolor{58.13} 2.43     & \gcolor{64.11} 2.68  \\ \hline
            \enquote{references to cafés and \newline coffee culture} (18/17350)                   & \gcolor{61.72} 2.58 & \gcolor{27.51} 1.15     & \gcolor{43.3} 1.81     & \gcolor{100} 4.18     & \gcolor{80.14} 3.35 \\ \hline
            \enquote{references to coffee and \newline its cultural significance} (18/28590)       & \gcolor{0} 0      & \gcolor{40.43} 1.69     & \gcolor{7.89} 0.33     & \gcolor{22.97} 0.96     & \gcolor{26.08} 1.09 \\ \hline
        \end{tabular}

        \caption{Activation of features (summed over all input terms and normalized for each column) based on inputs from the realm of these features. A comprehensible distinction between features cannot be identified (highest values should be on the diagonal). Feature (18/9463) is the most universally activated on coffee-related terms.}
        \label{tab:ConfusionMatrixLayer18Coffee}
    \end{table}
    
Table~\ref{tab:ConfusionMatrixLayer18Coffee} shows that feature descriptions seem to be an unfit tool to discriminate between features. We see activation of all features on all categories in general without any significantly higher activation of a feature on the terms that should cover the semantic region given by the category descriptions. 
This activation on all coffee-related terms seems to suggest that these features actually belong to the same semantically related cluster -- and were split by the fine-graining process described above. Solely the feature \enquote{references to drinks, particularly wine and coffee} (18/7546) has very low activations overall. We will take a closer look at this feature in the next Section \ref{subsec:ContextDependence}.
Contrary to our initial expectation, however, we cannot discriminate the five coffee-related features based on their activation on word-categories that were specifically chosen to fit their descriptions. This raises the question of why they are distinctly named and how to choose the best feature for steering in a specific case based on its name.

\subsubsection{Similar Features in Different Layers}\label{subsubsec:SimilarityDifferentLayer}
Since similar features can be identified within a single layer, might this trend also be observable across multiple layers? It's important to note that we can't expect to find the same features on every layer. Each SAE is trained on a single layer -- or, more precisely, a single point of the residual stream -- and there is nothing to ensure continuity of features between SAEs on different layers. 

Across all 32 layers of Llama 3.1 we found over $150$ features mentioning \enquote{coffee} (see Appendix~\ref{app:Features_Coffee}). While most of them have slightly different descriptions, we found three coffee-features across the network with the exact same description (see table~\ref{tab:references_to_coffee_features}). As discussed earlier, feature (29/30028) is a case of a poorly auto-interpreted feature, because it activates not on \enquote{coffee} and related terms, but on the token \enquote{coff}.

    \begin{table}[ht!]
        \small     
        \captionsetup{font=small} 
        \centering
        \begin{tabular}{lll}
        \hline
            Layer & Feature & Description \\ \hline
            6 & 25623 &  references to coffee \\
            19 & 12587 & references to coffee \\ 
            29 & 30028 &  references to coffee \\ \hline
        \end{tabular}
        \caption{Features that have the exact same description: \enquote{references to coffee}.}
        \label{tab:references_to_coffee_features}
    \end{table}

Although the three features have the same description, they exhibit different activation patterns on a coffee-related text (see Appendix \ref{app:Activation_referencesToCoffee}). They also influence the model differently when used for steering, as shown in Appendix \ref{app:Steering_Coffee_all} (see the steering differences for the five coffee-related features on the same layer 18 in Appendix \ref{tab:clamping_coffee_layer18_positive} and \ref{tab:clamping_coffee_layer18_negative} for comparison).
This raises a significant challenge for MI: What are we supposed to make of features that behave differently despite having the same description? Such features undermine the basic goal of a robust mapping between human interpretable features and model features. Anthropic is working on a method, called \textit{Sparse Crosscoder}, to enable such an analysis of persistent features throughout the residual stream of the model \citep[]{Crosscoder.2024}, but apart from their findings being preliminary at the time of writing, this just extracts one interpretable feature across layers. It does not ensure against features with the same interpretation but different behavior.

Our findings indicate that one cannot rely on the descriptions of features (that are based on their activation pattern on the input), either in a single layer or across different layers. In both cases, it seems to be more helpful to only analyze their steering suitability to draw conclusions about their capabilities.


\subsection{How Sensitive Are the Features to Context?}\label{subsec:ContextDependence}
As the examples of successful and unsuccessful feature interpretations in Section~\ref{subsec:scope} showed, feature quality and validity of the automated interpretations can differ significantly on a case by case basis. 
We would expect the feature \enquote{references to exposure to hazards or influential experiences} (see above), to activate only in a context that somehow refers to hazards or influential experiences, and not to all mentions of \enquote{exposure} in every context. This naturally leads us to ask about the context-dependence of features. In particular, we tested if there is a difference in activation on words with and without the context around them. We found examples of (1) features that are overly context-dependent, (2) features that have a high context-dependence, and (3) features that show little to no context-dependence whatsoever. 

In table \ref{tab:context_coffee_features}, we depict the context-dependence of three features: \enquote{references to drinks, particularly wine and coffee} (18/7546), \enquote{topics related to coffee, including its origins, effects, and cultural significance} (17/17320), and \enquote{mentions of coffee and related terms} (18/9463).
Note that we did not consider longer passages of context. In each case, we tested either no context at all, or just a few words preceding (or following) the word of interest. 

As we show, \enquote{references to drinks, particularly wine and coffee} (18/7546) is highly context-dependent, activating mainly on tokens following the phrases \enquote{(hot) cup of} or \enquote{drank}. This behavior is not even restricted to \enquote{wine and coffee} as the interpretation suggests, but also activates for \enquote{cup of poison} or \enquote{drank poison}. The full list of phrases can be found in Appendix \ref{app:Context_CoffeeWine}.
The right context appears to be essential to this feature, as the interpretation would suggest that it should activate on \enquote{coffee} on its own, but it does not.

\begin{table}[ht!]
    \renewcommand{\arraystretch}{1.4}
    \small     
    \captionsetup{font=small}
    \centering
    \begin{tabular}{|>{\raggedright}p{.3\textwidth} | >{\raggedright}p{.3\textwidth} | >{\raggedright\arraybackslash}p{.3\textwidth}|}
        \hline
        \textbf{(18/7546): references to drinks (wine, coffee)} & \textbf{(17/17320): related to coffee (origins, effects, cultural significance)} & \textbf{(18/9463): coffee and related terms} \\ \hline \hline
        coffee                                                                          & \redhighlight{red!17.05}{\strut coffee}                                                                               & \redhighlight{red!93.61}{\strut coffee} \\
        poison                                                                          & caffe\redhighlight{red!16.38}{\strut ine}                                                                             & Es\redhighlight{red!57.52}{\strut presso} \\
        cup \redhighlight{red!58.7}{\strut of} \redhighlight{red!48.55}{\strut coffee}  & \redhighlight{red!17.05}{\strut coffee} \redhighlight{red!100}{\strut from} \redhighlight{red!59.47}{\strut Africa}   & \redhighlight{red!47.37}{\strut caffe}\redhighlight{red!35.71}{\strut ine} \\
        cup \redhighlight{red!58.7}{\strut of} \redhighlight{red!40.1}{\strut poison} & zebras from Africa & C\redhighlight{red!51.5}{\strut off}\redhighlight{red!5.87}{\strut in} \\
        We \redhighlight{red!77.29}{\strut drank} \redhighlight{red!100}{\strut coffee}\redhighlight{red!52.17}{\strut .} & kwentoeler welrl newr \redhighlight{red!15.81}{\strut caffeine}                     & We drank our \redhighlight{red!71.05}{\strut coffee} \redhighlight{red!2.8}{\strut in} the morning \\
        We \redhighlight{red!77.29}{\strut drank} \redhighlight{red!56.52}{\strut poison}\redhighlight{red!52.42}{\strut .} & \redhighlight{red!23.39}{\strut Coffee}\redhighlight{red!74.24}{\strut :} \redhighlight{red!35.61}{\strut America}\redhighlight{red!66.67}{\strut 's} \redhighlight{red!83.71}{\strut most} \redhighlight{red!70.08}{\strut popular} \redhighlight{red!14.3}{\strut equipment} \redhighlight{red!71.59}{\strut for} & iwern owk ail \redhighlight{red!89.47}{\strut coffee} \\
        \hline
    \end{tabular}
    \caption{Context-dependence of features. \enquote{references to drinks, particularly wine and coffee} (18/7546) activates on tokens following the phrases \enquote{cup of} or \enquote{drank}. However, it does not activate on the term \enquote{coffee} in isolation. \enquote{topics related to coffee, including its origins, effects, and cultural significance} (17/17320) activates on tokens only in the surroundings of coffee-related terms. \enquote{mentions of coffee and related terms} (18/9463) activates on the token \enquote{coffee} (and related terms) regardless of the context.}
    \label{tab:context_coffee_features}
\end{table}

Second, \enquote{topics related to coffee, including its origins, effects, and cultural significance} (17/17320) is also context-dependent, but in a much more predictable way than the previous feature. While it lightly activates on words and tokens such as \enquote{coffee} or \enquote{caffeine}, we observe the strongest activations when there is a reference to coffee in its surrounding. For example, while there is no activation for \enquote{Africa} in connection with \enquote{zebras}, we can observe an activation for \enquote{Africa} in connection with \enquote{coffee}. See Appendix \ref{app:Context_CoffeeOrigin} for the full list of phrases and activations.

Lastly, the feature \enquote{mentions of coffee and related terms} (18/9463) does not seem to be very context-dependent at all. It activates on \enquote{coffee} and related terms (including tokens that make up these terms), regardless of the context. See Appendix \ref{app:Context_CoffeeRelated} for the full list.

All together, features seem to have varying degrees of context sensitivity that is not in any way apparent from their interpretation. Our first example (18/7546) is too context sensitive to be humanly interpretable at least with its current description. The second (17/17320) and third features (18/9463) are less context sensitive or only sensitive in a way one would expect for the description. Both are therefore much more interpretable and behave in expected ways. As we showed above (Section \ref{sec:replication}), the third feature (18/9463) can be very useful for steering.

\subsection{Features With Spurious Activations}\label{subsec:CuriousActivation}
In the course of our analysis, we encountered several features with activation patterns that clearly do not indicate semantic content. 

First, we found a family of features that have an activation density of over $90\%$, which we will refer to as \textit{hyperactive} features. Note that \citet[]{templeton2024scaling} mention the issue of \enquote{dead} features, i.e., features which were not active over a sample of $10^7$ tokens. As one might expect, larger SAEs produce larger numbers of dead features (in absolute, as well as relative terms) compared to their smaller counterparts. However, they do not mention features with an unusually high activation density. We give an overview of some of these hyperactive features -- features that activate on over $90\%$ of all tokens -- in table~\ref{tab:high-density_features} below. 

\begin{table}[ht!]
    \renewcommand{\arraystretch}{1.2}
    \small     
    \captionsetup{font=small} 
    \centering
    \begin{tabular}{p{.04\textwidth} p{.06\textwidth} p{.7\textwidth}p{.07\textwidth}} \hline
        Layer & Feature & Description & Density \\ \hline
        1 & 17701 & symbols and formatting elements commonly used in programming or markup languages & 93.80\%\\ 
        15 & 9478 & references to systems, mechanisms, or structures related to regulatory or operational frameworks & 93.50\%\\ 
        15 & 3179 & references to discussions or mentions of climate change and its impacts & 96.67\%\\
        24 & 29371 & numerical data and associated statistics related to various topics & 94.95\%\\ 
        30 & 24133 & references to accountability and oversight in institutional settings & 93.88\% \\ 
        7 & 27281 & financial and regulatory terms related to government policies and economic frameworks & 96.55\% \\ \hline
    \end{tabular}
    \caption{Features with an activation density higher than 90\%. In other words, they activate on more than 90\% of tokens.}
    \label{tab:high-density_features}
\end{table}

In some cases, one might expect such a relatively high activation density, such as with feature (1/17701), which is interpreted as \enquote{symbols and formatting elements commonly used in programming or markup languages}. In other cases, the explanation is not clear. For example, we do not expect a feature interpreted as \enquote{references to discussions or mentions of climate change and its impacts} (15/3179) to activate consistently on over $96\%$ of tokens.

Second, we found a family of features (table~\ref{tab:begin-of-text_activation_features}) that activate strongly on the \texttt{<|begin\_of\_text|>} token, a token that is only present at the very beginning of an input to signal the start of the token sequence. Their activation on the beginning of sequence can be orders of magnitude higher than their activation on normal text tokens. 
That (some) features activate on this beginning of sequence is to be expected -- all text samples used for training and description generation include this token. For example, our coffee feature also activates on the \texttt{<|begin\_of\_text|>} token, however, its activation strength is comparable to coffee-related text inputs.

\begin{table}[ht!]
    \renewcommand{\arraystretch}{1.2}
    \small     
    \captionsetup{font=small} 
    \centering
    \begin{tabular}{p{.04\textwidth} p{.06\textwidth} p{.67\textwidth}p{.1\textwidth}}
    \hline
        Layer & Feature & Description & Activation \\ \hline
        1 & 5371 & occurrences of the word "this" & 245.00\\ 
        1 & 26183 & HTML or CSS structures and elements & 244.00\\ 
        10 & 11036 & formatted sections or placeholders within text, possibly indicative of form fields or structured data & 140.00 \\ 
        10 & 19323 & numerical data related to years and events in history & 123.50 \\ 
        20 & 6246 & monetary values and financial information & 156.00 \\ 
        20 & 21737 & phrases related to weight loss and body shaping techniques & 100.50 \\ 
        30 & 19188 & phrases and concepts related to guidance and support in educational contexts & 140.00 \\ 
        30 & 20145 & topics related to skincare and personal well-being & 131.00 \\ \hline
    \end{tabular}
    \caption{Features activating on \texttt{<|begin\_of\_text|>} token. Their listed maximum activation is only present for the \texttt{<|begin\_of\_text|>} token.}
    \label{tab:begin-of-text_activation_features}
\end{table}

Despite this high activation value on the beginning of a text, we observe that the model output can be steered consistently with the interpretation (that does not refer to the beginning of sequence) in many cases.
We show this using the features \enquote{monetary values and financial information} (20/6246, \texttt{<|begin\_of\_text|>} activation: 156.0) and \enquote{phrases related to weight loss and body shaping techniques} (20/21737, \texttt{<|begin\_of\_text|>} activation: 100.5) as examples (see Appendix \ref{app:ContextDependence} for the results). In comparison to other features, however, the model behavior seems to be a little bit more erratic and hard to control.

It is not entirely clear how steering in general should or could influence a token that is by design only present at the very beginning of a sequence. There are, however, other special tokens at work during training and deployment of LLMs that are not only present at the beginning. Dialogue tokens differentiate what the user said and what the model generated. Masking tokens are used to train a model on filling gaps with the right tokens. And some special tokens differentiate instruction from conversation for a model. An unusual response to these tokens by the SAE might have unexpected consequences when steering the model. 


\subsection{Can Features be Steered Precisely?}\label{subsec:Coherence}
We showed above that, generally speaking, it is possible to steer model behavior by attenuating or amplifying the activation pattern associated with certain features. Are the effects of feature steering on model behavior always meaningful and consistent? 
As we will show in the following, the quantitative aspects of model steering can be rather nuanced and fickle. 

\bigskip\noindent\textbf{(1) \enquote{mentions of tennis and related terms} (16/12285)}\smallskip\\
The only tennis-related feature in layer 16 -- \enquote{mentions of tennis and related terms} (16/12285) -- is a good example of how steering may not have the effect one expects. As we depict in table~\ref{tab:clamping_1-2_tennis}, the input in this case is \enquote{My favorite sport is} and the baseline model names baseball as its standard output. Clamping the tennis-related feature by a coefficient of $1$ changes the output of the model from baseball to soccer. Furthermore, while the output of the baseline model consistently mentions baseball over two paragraphs, the steered model also mentions basketball -- but no mention of tennis. For a steering coefficient of $2$, the model output then does discuss tennis, famous tennis players, and their achievements. In a second paragraph, however, the model continues mentioning a variety of other sports. 

It remains unclear what explains this unexpected behavior. It could be that the interpretation is insufficient, or that steering a feature can sometimes involve unexpected detours. While tennis content is generated for a steering coefficient of $2$, one would not expect it to go via soccer for a steering coefficient of 1. 

\begin{table}[ht!]        
    \small     
    \captionsetup{font=small}     
    \centering          
    \begin{tabular}{|p{.12\textwidth}|p{.82\textwidth}|}\hline
        Steering & My favorite sport is... \\ \hline\hline
        Baseline & ... \textbf{baseball}. I like to play \textbf{baseball} and watch it on TV. I also like to go to the games at the ballpark. My favorite team is the Minnesota Twins.\newline I have been playing \textbf{baseball} since I was 5 years old. When I was 5, my dad signed me up for Little League Baseball. He didn't know how to play but [...]\\
        Coeff $1$ & ... \textbf{soccer}. I like to play it and watch it. I play \textbf{soccer} with my friends in the park near my home every weekend. I like to watch it on TV when there are matches between famous teams.\newline I like to play \textbf{basketball}, too, but I don't play it very often. My friends and I sometimes play \textbf{basketball} in the park after we [...] \\
        Coeff $2$ & ... \textbf{tennis}. I like to play tennis and watch it on TV. I'm a big fan of \textbf{Serena Williams and Roger Federer}. They are the best players in the world and they have won many Grand Slam titles.\newline I also like to play other sports, such as golf, swimming, and skiing. I enjoy watching \textbf{football} and \textbf{basketball} as well.\newline What [...] \\ \hline
    \end{tabular} 
    \caption{Steering capability of feature \enquote{mentions of tennis and related terms} (16/12285). Positively clamping this feature by $1$ changes the mentioned sport from \enquote{baseball} to \enquote{soccer} or \enquote{basketball}. Increasing the factor further to $2$ generates a description of \enquote{tennis}, before mentioning other sports.} 
    \label{tab:clamping_1-2_tennis} 
\end{table}

\bigskip\noindent\textbf{(2) \enquote{references to the effects of coffee on health} (15/4922)}\smallskip\\
This is one of the many coffee-related features we found. As depicted in Appendix \ref{app:CoherenceSteering_coffeehealth}, we provided the model with the input \enquote{My favorite drink is}. With a steering coefficient of $5$, this feature works similarly to other coffee features we found and studied: it changes the model output from \enquote{margarita} to \enquote{coffee}. For slightly higher ($6$ and $8$) steering coefficients, however, the model generated \enquote{a simple cup of hot tea} as its first output, only afterwards mentioning coffee. In this case the switch to tea seems to be a first sign of breaking, as a steering coefficient of $8$ starts to affect the correctness of the syntax of the output.

\bigskip\noindent\textbf{(3) \enquote{references to pigeons and airborne topics} (18/16371)}\smallskip\\
This feature has a rather peculiar interpretation. It combines a  specific type of bird -- \enquote{pigeons} -- with the general addition \enquote{and airborne topics}. Considering only its interpretation, it is not obvious how the model behavior would be affected, but an increased tendency to mention pigeons seems likely. This is especially true, as the feature also activates on \enquote{Racing Pigeon}.
As can be seen in Appendix \ref{app:CoherenceSteering_pigeons}, we provided the model with the input \enquote{An example for a bird is}, and the answer of the baseline model is \enquote{hummingbird}. Contrary to our initial expectations, the steered model does not generate \enquote{pigeon} as its output, but \enquote{American Robin}. This output is stable for increasing steering coefficients ($2$, $5$, and $10$), with only the additional information provided by the model changing -- being correct descriptions for American Robins.

\section{Discussion}\label{sec:Analysis}
\subsection{Reproducibility and Scope of Interpretable Features}
Our investigation successfully reproduced Anthropic's core findings, even while using a different model. We confirmed that SAEs can extract semantically meaningful features from LLMs, and that these features can be manipulated to produce syntactically coherent text that emphasizes or de-emphasizes particular concepts (see Section \ref{sec:replication} and \ref{subsubsec:GoodFeatures}).

However, this successful reproduction comes with significant caveats. Alongside interpretable features, our analysis revealed numerous features whose automated labels fail to align with their actual behavior during activation and steering. We showed this behavior in selected cases with features like \enquote{references to exposure to hazards or influential experiences} (18/14711) and \enquote{the word \enquote{Morocco} and its variations or related concepts} (18/6893) in Section \ref{subsubsec:BadFeatures}. Importantly, we saw that there is no consistent way of determining if a feature behaves as expected just by looking at its description. This observation raises concerns about the completeness of existing presentations of SAE results, which tend to showcase the most successful and interpretable features while leaving the proportion of less interpretable features unclear.

The challenge of selective presentation in interpretability research has been previously acknowledged in the literature. \citet[]{Rauker.2023} explicitly discuss the validity of interpretations and the risks of cherry-picking results. When companies conduct research on proprietary models that cannot be independently audited, they bear enhanced responsibility for comprehensive and honest reporting. Selective presentation may advance a compelling narrative about MI, but it does not provide the evidence base needed to assess whether these techniques can actually control model behavior reliably enough for real-world deployment. Notably, \citet[]{marks2025} originally claimed to offer demonstrations on \enquote{non-cherry-picked tasks of practical value}, but this language was removed in their 2025 update, suggesting evolving recognition of these concerns within the field. Our investigation shows just how easy it is to find examples that undermine the goals of MI and how hard it is to ensure against them. 

Associated with this, another critical yet often implicit assumption in SAE-based interpretability research concerns the relationship between identified features and human concepts. The intuitive appeal of interpretable features stems partly from the suggestion that they correspond to concepts humans recognize and employ. However, this assumption warrants careful scrutiny.

The question is not whether all features learned by language models are \enquote{bad} or spurious -- transfer learning demonstrates that neural networks learn generalizable patterns beyond task-specific artifacts \citep[282]{Freiesleben.2026}. Rather, the issue concerns which features correspond to human concepts and how we would determine this correspondence. As Freiesleben notes, research on adversarial examples and robustness suggests that neural networks learn both, human and non-human concepts \citep[283-284]{Freiesleben.2026}. The nature of these non-human concepts and whether humans can in principle understand them remains an open question.

Our analysis suggests that the features highlighted in prominent interpretability demonstrations may not be representative of the broader feature distribution. While some features clearly activate in patterns humans recognize and can be described using familiar concepts, many others resist straightforward interpretation or exhibit activation patterns that do not map cleanly onto human conceptual categories. The existence of interpretable features does not imply that most or all features share this property.

This observation has implications for how we evaluate interpretability claims. Anthropic's earlier blog posts sometimes suggest broad interpretability of identified features, though their later work on feature steering adopts more cautious language about feature labels and their reliability. The shift in framing may reflect growing recognition that human interpretability of a feature -- our ability to assign it a descriptive label -- does not necessarily mean the feature corresponds to a concept humans employ in their own reasoning, or that our interpretation captures the feature's actual computational role.

To make claims about human concepts in language models genuinely worthwhile for interpretability, we need systematic understanding of when human-recognizable concepts emerge and verification that they \enquote{[...] represent in the sense of coactivating AND taking the functional role of concepts.} \citep[288]{Freiesleben.2026} This requires demonstrating not only that features can be described in human terms, but that they function computationally in ways analogous to how humans use those concepts.

\subsection{Interpretation-Behavior Gap}
A central finding of our analysis concerns the reliability of feature interpretations. We identified disconnects between the automated descriptions of features on the one hand, and their activation patterns across text samples and their effects when used for model steering on the other hand. This misalignment poses significant challenges for claims about feature interpretability.

The automated labeling methods employed to describe features -- which typically identify topics in text samples where features activate -- do not necessarily capture the directionality or causal role of those features. As \citet[]{Durmus.2024} note in their work on feature steering, labels related to concepts like discrimination may affect discrimination-related outputs \enquote{[...] but not necessarily increase (or decrease) discrimination or bias in predictable ways.} Our own experiments confirm this concern: (1) we observed cases where features seemingly activate in contexts consistent with their labels but produce unexpected or incoherent effects when used for steering, e.g., the feature \enquote{references to pigeons and airborne topics} (18/16371) in Section \ref{subsec:Coherence}; (2) there exist features like \enquote{references to coffee} (29/30028) where activation and steering consistently influence the token \enquote{coff}, but whose automated interpretation suggests a conceptual connection to coffee (see Section \ref{subsubsec:BadFeatures}); (3) even though we find features with the exact same description, namely \enquote{references to coffee}, across different layers, these do not exhibit the same behavior for neither activation nor steering (see Section \ref{subsubsec:SimilarityDifferentLayer}).

This interpretation-behavior gap appears to have multiple sources. First, feature activation exhibits context-dependence that complicates straightforward interpretation. What appears to be a coherent concept may in some cases reduce to activation on particular words or tokens in specific contexts, creating an illusion of more abstract conceptual representation (see the context-dependent features in Section \ref{subsec:ContextDependence}). Second, features with highly similar descriptions often coexist within the same layer (see the five coffee-related features in layer 18 shown in Section \ref{subsubsec:SimilaritySingleLayer}), raising questions about their functional differentiation and relationship to one another. Without clear principles for distinguishing between related features, it becomes difficult to claim precise understanding of what any individual feature represents.

In addition, we found two types of activation patterns that appear entirely spurious (see Section \ref{subsec:CuriousActivation}), namely features with a very high activation density, as well as features whose activation is disproportionally high for the \texttt{<|begin\_of\_text|>} token. Both of these activations are not detectable in the feature descriptions themselves. This points to another type of gap between interpretation and behavior. As it is not clear where their behavior comes from, features seem to be able to encode particular behavior that is not encodeable in their interpretation.

\subsection{Fragility of Feature Steering}
Our experiments reveal that feature steering is a highly sensitive operation characterized by several forms of fragility. The effectiveness and meaning of steering depends 
on the quantitative magnitude of the steering coefficient, the accuracy of the feature's interpretation, and whether the feature responds as expected in diverse contexts. We present examples of malfunctioning steering -- e.g., the steering of feature \enquote{mentions of tennis and related terms} (16/12285) resulting in text on sports other than tennis -- in Section \ref{subsec:Coherence}.

The magnitude of steering interventions requires careful calibration, and in general depends on the specific layer and network depth. See, e.g., the feature \enquote{references to the effects of coffee on health} (15/4922) in Section \ref{subsec:Coherence}: the steering works well for a factor of 5, produces unexpected results for 6, and breaks the model output for a factor of 8. Other features only start to exhibit an influence on model behavior with a steering coefficient of 10 and more (see the lists of clamped coffee-features in Appendix \ref{app:Steering_Coffee_all}). \citet[]{Durmus.2024} acknowledge this fact, but leave questions unanswered (see Section \ref{subsec:steering}). Small changes in steering strength can produce disproportionate effects, and steering for one intended concept may inadvertently affect homonyms, semantically similar terms, or unrelated linguistic patterns (see the examples of mis-interpreted features in Section \ref{subsubsec:BadFeatures}). The boundaries of what is being manipulated remain unclear, complicating efforts to achieve targeted behavioral modifications.

Due to the layer-specificity of SAEs, our approach does not allow to draw any final conclusions on the layer-dependence of features. Because SAEs are trained on individual layers, one cannot be sure if two identically interpreted features in subsequent layers actually represent the same concept. However, we encountered several nearly identically interpreted features in a single layer (see the coffee-related features in Section \ref{subsubsec:SimilaritySingleLayer}), which begs the question how these can be meaningfully distinguished and then used for steering purposes. While these findings are only (multiple) instances, they significantly undermine the suggested generality and applicability of this route towards mechanistic interpretability. The broad expectation that in general we will be able to find well-interpreted features across a network that can be reliably steered, is not warranted.

From an AI safety perspective, this fragility is particularly concerning. Any attempt to use SAE-based steering as a control mechanism would need to account for sensitivity to layer positioning, precise parameterization, interpretive accuracy, and potential side effects -- a combination of requirements that suggests limited robustness for practical deployment.

\subsection{Representational Faithfulness and Reasoning}
Beyond questions of interpretability and steering reliability, it is an open question whether the features produced by SAEs provide faithful abstractions of the language model's actual \enquote{reasoning} or inference processes. While our approach does not allow us to answer this question systematically, we believe the following two issues to be central if one wants to make progress in the direction of these strong representational claims.

First, one would need to show coherence between thematically similar concepts across different layers. If SAE-features captured fundamental reasoning primitives used throughout the model's computation, we would expect to find systematic relationships or consistent representations of related concepts as information flows through the network. The absence of such coherence suggests that features may capture layer-specific activation patterns rather than stable cognitive building blocks. Note, however, that this might be due to the training of SAEs on individual layers of the network. The utility of generalized approaches -- such as crosscoders \citep[]{Crosscoder.2024} -- in this context is still up to debate.  

Second, one would need to find feature integration or composition that mirrors standard forms of reasoning. If the identified features genuinely represented the model's conceptual vocabulary, we would expect to observe systematic interactions between features during inference -- compositions and combinations that reflect logical relationships or inferential steps. 
An initial move toward that end is taken by \citet{templeton2024scaling} when discussing the potential role of features as computational intermediates. They try to determine whether the features are useful for examining the intermediate computational steps that would be necessary for a model to reach a multi-step conclusion. As they show, features like \enquote{Kobe Bryant}, \enquote{California}, \enquote{Capitals}, \enquote{Los Angeles}, and \enquote{Los Angeles Lakers} are evoked by the model to complete the sentence \enquote{The capital of the state where Kobe Bryant played basketball is}. However, there is still a long way to go before reaching reasoning. The authors acknowledge that there are only a few examples where these intermediate features are this well interpretable. Furthermore, they can only demonstrate activation of these features, not any further interaction. Therefore, this does not definitively answer the question of whether these features truly capture the model's representational primitives or merely correlate with particular activation patterns.

The risk, as \citet[289]{Freiesleben.2026} cautions, is that \enquote{[p]eople want to believe that ANNs are logical concept machines that only need to be scrutinized to become fully transparent.} This narrative, if accepted uncritically, can lead to research prone to confirmation bias -- focusing on impressive examples of interpretable features while underestimating the proportion of features that resist interpretation or the challenges in verifying that interpretations reflect actual computational function. Such bias has real consequences for users who may overestimate our ability to understand and control model behavior based on selective demonstrations.

\section{Conclusion} \label{sec:Conclusion}
Mechanistic interpretability, as a form of explainable AI, refers to a broad set of recent approaches to understanding, and ultimately controlling, the behavior of LLMs. The core idea is to construct semantic representations -- so-called \enquote{features} -- from the neural activation patterns of the network, that are both human-interpretable and controllable. There are different approaches to discover, extract, and study such functionally useful features in LLMs. One particular technique used in MI research are sparse autoencoders. In this work, we analyzed the work on SAE-based extraction of interpretable features conducted by Anthropic. 

Our replication in Section~\ref{sec:replication} confirmed that interpretable features exist and activate appropriately in expected contexts. Our testing revealed that features show genuine specificity, with activation strength correlating with contextual relevance. We also confirmed that features can effectively steer model behavior in predictable ways, with steering effectiveness relative to layer position and size of the steering coefficient.

In Section~\ref{sec:investigation}, we critically examined the potential to generalize specific findings in the context of SAE-based features to larger contexts. This has important implications for the role of MI research to provide model understanding, and to be used as a tool for AI safety.

The interpretability paradigm emphasizes understanding internal mechanisms by identifying what features represent and how they activate in response to inputs. While this theoretical project remains valuable for researchers seeking to understand neural network computations, it may not be the most relevant framework for addressing practical safety and deployment concerns. For broader societal purposes, the critical question is not whether we can interpret a model's internal representations, but whether we can reliably control its output behavior. \citet[]{London.2019} argues that when causal knowledge is incomplete, careful empirical validation of accuracy and reliability should take priority over demands for explanation in medical decision making. The same logic applies to LLMs that affect millions of users on a daily basis: If the goal is AI safety, reliable control of the model's output behavior is more important than detailed mechanistic interpretations. 

One step in that direction is taken by \citet{templeton2024scaling} when discussing the features as computational intermediates (see the analysis of the sentence \enquote{The capital of the state where Kobe Bryant played basketball is} above). In the course of this analysis, they discuss \textit{ablation} and its local approximation \textit{attribution}. Ablation is the gold standard when it comes to determining the influence of a model part -- in this case a feature -- on the output. To measure ablation, each feature one after another is turned off (set to zero) for each predicted token in the sequence to measure the feature's full (possibly non-linear) contribution to that prediction. While this is a precise measurement, it requires significant computational resources. \citet{templeton2024scaling} thus use a form of attribution patching to approximate the ablation measurements. Simply put, they approximate the influence of a feature on a given prediction by its decoder weight. While this method is not able to measure non-linear effects, it saves computational cost, as it does not require additional forward-passes through the model for each feature. In a footnote they then show that the correlation between ablation and attribution is much higher than the correlation between ablation and activation \citep{templeton2024scaling}. This means that the activation pattern associated with a feature, which is computed based on the input to a model, is not a great way to predict its ability to influence output behavior! However, as we discussed above, activation-first is the way features are currently approached.

This suggests a necessary shift in focus: from the input-to-feature relationship toward the feature-to-output relationship. Rather than asking primarily \enquote{what does this feature represent?} we should prioritize \enquote{what behavioral effects does manipulating this feature produce, and how reliably can we predict those effects?} The distinction parallels pharmaceutical development: we do not require complete mechanistic understanding of how a drug affects every molecular pathway in the body before approving it for use. Instead, we demand rigorous evidence about its effects on patient outcomes through systematic testing across diverse populations, clear documentation of efficacy and side effects, and reproducible protocols for safe administration. Similarly, for AI safety, the question is not whether we fully understand the internal computations, but whether we can reliably predict and control what the system will output under various conditions.

\backmatter 

\newpage

\setlength{\textfloatsep}{0pt}
\setlength{\intextsep}{0pt}
\setlength{\floatsep}{0pt}

\begin{appendices}\label{sec:appendix}

\section{Details: Replicating Anthropic's Results}

\subsection{Setup Details}\label{app:SetupDetails}
\begin{table}[ht!]
    \renewcommand{\arraystretch}{1.2}
    \small     
    \captionsetup{font=small} 
    \centering
    \begin{tabular}{|p{.25\textwidth}|p{.65\textwidth}|} \hline
        \textbf{LLM}            & ~ \\
        Name           & Llama 3.1\\
        Parameters     & 8 Billion \\ \hline
        \textbf{SAE}            & ~ \\
        Description  & Llama Scope (residual stream) \\
        Trained SAEs & Llama3\_1-8B-Base-L[LAYER]R-8x (\hyperlink{https://huggingface.co/OpenMOSS-Team/Llama3_1-8B-Base-LXR-8x/tree/main}{Hugging Face}) \\
        Factor       & 8x (Size of hidden layer x 8 = 32,768 features) \\ \hline
        \textbf{Interpretation} & ~ \\
        Model          & o4-mini \\
        Dataset        & cerebras/SlimPajama-627B (\hyperlink{https://huggingface.co/datasets/cerebras/SlimPajama-627B}{Hugging Face}) \\ \hline
        \textbf{Steering}       & ~ \\
        Temperature    & 0.5 \\
        No. of output Tokens generated      & 70 \\
        Frequency penalty   & 1.0 \\
        Seed           & 16 \\
        Strength Multiplier & 1 \\
        ~ & Strength Multiplier of 1: Steering strengths given in the paper are identical with the steering coefficients that are multiplied with the feature vectors, and are, therefore, a direct measurement of the influence on the LLM.\\ \hline
    \end{tabular}
    \caption{Parameters for SAEs in the paper.}
    \label{tab:parameters}
\end{table}
\subsection{Replicating Feature Activation Specificity}\label{app:AnthropicCategoriesSentences}
Sentences in for each of Anthropic's specificity categories to test activation of feature \enquote{mentions of coffee and related terms} (18/9463):
\begin{itemize}
    \item Category 0: The feature is completely irrelevant throughout the context
    \begin{itemize}
        \item The quantum fluctuations in vacuum energy remain a mystery in theoretical physics.
        \item She spent the afternoon learning about ancient Mesopotamian irrigation techniques.
        \item Satellite imagery revealed unexpected changes in polar ice caps over the past decade.
        \item He carefully cataloged every species of beetle he discovered in the rainforest expedition.
    \end{itemize}
    \item Category 1: The feature is only vaguely related
    \begin{itemize}
        \item The café down the street serves pastries and teas in addition to their other offerings.
        \item Early morning routines often involve some kind of beverage before heading to work.
        \item Many people gather in social spaces like bistros and lounges to chat before work.
        \item She walked past a bakery that also sold drinks in takeaway cups on her commute.
    \end{itemize}
    \item Category 2: The feature is only loosely related to the highlighted text
    \begin{itemize}
        \item Although she preferred herbal tea, she occasionally grabbed a latte when feeling tired.
        \item He thought about brewing something warm like espresso while reading the newspaper.
        \item Sometimes she opted for a macchiato instead of her usual hot chocolate.
        \item He debated ordering a frappé as he scanned the drink menu at the counter.
    \end{itemize}
    \item Category 3: The feature cleanly identifies the activating text
    \begin{itemize}
        \item She poured herself a steaming mug of dark-roast coffee to jump-start her day.
        \item The rich aroma of freshly ground Arabica beans filled the air as the barista tamped the espresso shot.
        \item He sipped a creamy cappuccino topped with frothed milk at the busy coffeehouse.
        \item The cold brew coffee dripped slowly into the carafe, promising a bold and smooth flavor.
    \end{itemize}
\end{itemize}

\section{Details: Do the Findings Generalize?}\label{app:DetailedInvestigation}
\subsection{Examples of Successful Interpretations}\label{app:GoodFeatures}
\subsubsection{\enquote{expressions of regret, mourning, and sadness} (17/16706)}\label{app:GoodFeatures_regret}
\noindent
\textbf{Activation pattern:}\\
\begin{table}[ht!]
    \small     
    \captionsetup{font=small}     
    \centering     
    \begin{tabular}{p{.96\textwidth}} \hline
        The weight of \redhighlight{red!100}{\strut regret} settled like a shadow over \redhighlight{red!19.87}{\strut her} \redhighlight{red!26.03}{\strut heart}\redhighlight{red!8.72}{\strut ,} each memory a \redhighlight{red!8.4}{\strut painful} echo \redhighlight{red!25}{\strut of} choices \redhighlight{red!8.78}{\strut she} could not undo. G\redhighlight{red!76.41}{\strut rief} clung to her like a second skin, and the silence of \redhighlight{red!17.56}{\strut mourning} filled the empty spaces where laughter used to live. She moved through her days with a quiet \redhighlight{red!30.51}{\strut ache}, burdened \redhighlight{red!21.79}{\strut by} \redhighlight{red!82.56}{\strut sorrow} and \redhighlight{red!7.95}{\strut the} haunting presence of what was lost. Rem\redhighlight{red!60.77}{\strut orse} whispered in every still moment, reminding her of the paths not taken and the words left unsaid. In the quiet corners of her mind, \redhighlight{red!36.41}{\strut sadness} bloomed like wilted flowers—tender, a\redhighlight{red!21.67}{\strut ching}, and impossible to forget.\\ \hline
    \end{tabular}
    \caption{Activation of feature \enquote{expressions of regret, mourning, and sadness} (17/16706) on English text. The feature related to regret, mourning, and sadness activates on and around these terms.}
    \label{tab:activation_sadness}
\end{table}

\newpage
\noindent
\textbf{Steering Capabilities:} \\
\begin{table}[ht!]     
    \small
    \captionsetup{font=small}
    \centering     
    \begin{tabular}{|p{.12\textwidth}|p{.82\textwidth}|}\hline
    Steering & Summer is the time of the year where I... \\ \hline\hline
    Baseline & ... \textbf{get to take a break} from school and work and \textbf{just hang out with my family}. We are very fortunate to live close to a lot of outdoor activities, including beaches, hiking trails, and nature parks.\newline \textbf{Our favorite place to go is the beach.} Growing up in Southern California, I have been exposed to so many beaches [...] \\
    Coeff $10$& ... \textbf{have to take my kids} to the beach, and I am not a beach person. \textbf{I don't like} the sand, and swimming in the sea. \textbf{So I really hope that one day, I could go to a beach with no sand.}\newline I am not the one who has to go there. My kids have to go [...]\\ \hline
    \end{tabular} 
    \caption{Steering capability of feature \enquote{expressions of regret, mourning, and sadness} (17/16706). Positively clamping this feature changes the output from an overall happy story on holiday activities to a narration full of regretful rhetoric.}
    \label{tab:clamping_10x_sadness-a}
\end{table}

\begin{table}[ht!]  
    \small
    \captionsetup{font=small}
    \centering
    \begin{tabular}{|p{.12\textwidth}|p{.82\textwidth}|}\hline
        Steering & The best thing that ever happened to me was... \\ \hline\hline
        Baseline & ... \textbf{the day I met you} and I want to spend the rest of my life with you. Let's make our love story last forever.\newline I have been waiting for someone like you my whole life.\textbf{ I am so happy that we found each other} and I can't wait to spend forever with you. Happy Anniversary!\newline You are [...]\\
        Coeff $10$ & ... \textbf{the day I lost my sight}. It was a rainy day and I could not see the rain. I was in a place where there was no one, but it was a place where there were no people.\newline It is true that losing your sight is \textbf{one of the most unfortunate things} that can happen to anyone, but it [...]\\ \hline
    \end{tabular} 
    \caption{Steering capability of feature \enquote{expressions of regret, mourning, and sadness} (17/16706). Positively clamping this feature changes the output from from an overall happy memory to a narration on permanently lost sight.} 
    \label{tab:clamping_10x_sadness-b}
\end{table}

\newpage
\subsubsection{\enquote{instances of the phrase \enquote{would have} or similar expressions indicating hypothetical or conditional situations} (19/12491)}\label{app:GoodFeatures_WouldHave}
\noindent
\textbf{Activation pattern:}\\
\begin{table}[ht!]     
    \small    
    \captionsetup{font=small}     
    \centering     
    \begin{tabular}{p{.96\textwidth}} \hline
        If things had gone differently, perhaps she would \redhighlight{red!18.53}{\strut be} standing in a brighter place, untouched by the weight of what could \redhighlight{red!29.91}{\strut have} been. Had she spoken up that day, maybe everything would \redhighlight{red!100}{\strut have} changed—perhaps the outcome would \redhighlight{red!82.14}{\strut have} bent toward something softer, something whole. Suppose she'd taken that chance, followed that fleeting instinct—would she now be free of doubt? What if the timing had been just a little kinder, the circumstances a little more forgiving? In another version of the story, maybe she isn't haunted by questions, living instead in the calm certainty of choices that worked out.\\ \hline
    \end{tabular}
    \caption{Activation of feature \enquote{instances of the phrase \enquote{would have} or similar expressions indicating hypothetical or conditional situations} (19/12491) on English text. The feature related to hypothetical or conditional situations activates on \enquote{have} following terms like \enquote{could} and \enquote{would}.}
    \label{tab:activation_wouldhave}
\end{table}

\noindent
\textbf{Steering Capabilities:}\\
\begin{table}[ht!]  
    \small
    \captionsetup{font=small}   
    \centering     
    \begin{tabular}{|p{.12\textwidth}|p{.82\textwidth}|}\hline
        Steering & I am going to... \\ \hline\hline
        Baseline & ... share my own experience of my first ever visit to a Psychic Medium. I have been to a few different ones in the past year and this is the one that really stood out for me.\newline I was recommended by a friend who had been to see this lady and really enjoyed it. I was sceptical as I have never [...]\\
        Coeff $10$ & ... \textbf{have to take a look} at that, I do not like the idea of having to go through the entire 3d model of the vehicle and making it look like it was in a wreck. I am sure there is an easier way if you can believe that.\newline I know there is a way to use the same mesh for multiple objects, \textbf{but} I [...]\\ \hline
    \end{tabular} 
    \caption{Steering capability of feature \enquote{instances of the phrase \enquote{would have} or similar expressions indicating hypothetical or conditional situations} (19/12491). Positively clamping this feature changes the output to contain a lot more conditional statements.} 
    \label{tab:clamping_15_wouldhave-a} 
\end{table}

\begin{table}[ht!]  
    \small
    \captionsetup{font=small}   
    \centering     
    \begin{tabular}{|p{.12\textwidth}|p{.82\textwidth}|}\hline
        Steering & Photosynthesis is characterized by... \\ \hline\hline
        Baseline & ... the ability of plants to use solar energy to convert carbon dioxide and water into carbohydrates. The carbohydrates are stored in the plant and can be used as a food source for humans and animals. Photosynthesis occurs in the chloroplasts of plant cells, which contain the chlorophyll that absorbs light energy from the sun.\newline The [...] \\
        Coeff $10$ & ... the ability of plants to use solar energy to convert carbon dioxide and water into carbohydrates. The carbohydrates are stored as starches in the plant, but they \textbf{would be} useless to us if they were not eventually broken down and released as sugars, if we wanted to have some of them. In addition, the plant \textbf{would have been} useless had it not been able [...] \\ \hline
    \end{tabular} 
    \caption{Steering capability of feature \enquote{instances of the phrase \enquote{would have} or similar expressions indicating hypothetical or conditional situations} (19/12491). Positively clamping this feature changes a scientific description to contain a lot of conditional phrases like \enquote{would have}.} 
    \label{tab:clamping_15_wouldhave-b} 
\end{table}

\newpage
\subsubsection{\enquote{references to lists and list-making processes} (16/24361)}\label{app:GoodFeatures_List}
\noindent
\textbf{Activation pattern:}\\
\begin{table}[ht!]     
    \small     
    \captionsetup{font=small}     
    \centering     
    \begin{tabular}{p{.96\textwidth}} \hline
        Before leaving, she double-checked her tasks to make sure nothing was overlooked. Her \redhighlight{red!85,23}{\strut checklist} included: 1) cancel the mail delivery, 2) set the security alarm, and 3) leave a spare key with the neighbor. On her desk, she left a \redhighlight{red!59,92}{\strut smaller} note with labeled \redhighlight{red!16,56}{\strut items} she couldn't forget to pack: A) her passport, B) phone charger, and C) the envelope marked *For When I'm Gone*. Each \redhighlight{red!100}{\strut list}\redhighlight{red!64,98}{\strut ,} \redhighlight{red!62,45}{\strut though} \redhighlight{red!72,57}{\strut simple}\redhighlight{red!60,76}{\strut ,} \redhighlight{red!53,59}{\strut held} significance \redhighlight{red!27,64}{\strut —} practical \redhighlight{red!14,45}{\strut on} \redhighlight{red!28,06}{\strut the} surface, but deeply tied to the quiet finality of her departure.\\ \hline
    \end{tabular}
    \caption{Activation of feature \enquote{references to lists and list-making processes} (16/24361) on English text. The feature related to lists activates terms like \enquote{list} and \enquote{item}, as well as on the m-dash.}
    \label{tab:activation_list}
\end{table}

\newpage
\noindent
\textbf{Steering Capabilities:}\\
\begin{table}[ht!]
    \small
    \captionsetup{font=small}
    \centering     
    \begin{tabular}{|p{.12\textwidth}|p{.82\textwidth}|}\hline
        Steering & Let us talk about... \\ \hline\hline
        Baseline & ... the best way to find a good and reliable web design company. There are many companies out there that claim to be the best, but how do you know which one is right for your business? \textbf{In this blog post, we will discuss some tips} on how to find a web design company that is perfect for your needs!\newline There are many factors to consider [...] \\
        Coeff $5$ & ... the best way to find a good and reliable web design company. There are many companies out there but not all of them will be able to give you what you really want. \textbf{For your convenience, here is a list} of tips that you can follow in order to find the right company.\newline \textbf{First,} you should make sure that the company has an updated website [...] \\ \hline
    \end{tabular} 
    \caption{Steering capability of feature \enquote{references to lists and list-making processes} (16/24361). Positively clamping this feature changes the output to contain the term \enquote{list}, as well as \enquote{First}, indicating an enumeration.} 
    \label{tab:clamping_5_listmaking-b} 
\end{table}
\begin{table}[ht!]
    \small
    \captionsetup{font=small}
    \centering     
    \begin{tabular}{|p{.12\textwidth}|p{.82\textwidth}|}\hline
        Steering & Photosynthesis is characterized by... \\ \hline\hline
        Baseline & ... the ability of plants to use solar energy to convert carbon dioxide and water into carbohydrates. The carbohydrates are stored in the plant and can be used as a food source for humans and animals. Photosynthesis occurs in the chloroplasts of plant cells, which contain the chlorophyll that absorbs light energy from the sun.\newline The [...] \\
        Coeff $15$ & ... the presence of a \textbf{list} of pigments, which are present in the thylakoid membrane. \textbf{The list} is as follows:\newline (a) Chlorophyll a\newline (b) Chlorophyll b\newline (c) Chlorophyll c\newline (d) Xanthophylls\newline (e) Carotenoids\newline (f) Phycoerythrin\\ \hline
    \end{tabular} 
    \caption{Steering capability of feature \enquote{references to lists and list-making processes} (16/24361). Positively clamping this feature changes a scientific description to contain a list of terms.} 
    \label{tab:clamping_15_listmaking-a} 
\end{table}

\newpage
\subsection{Examples of Mis-interpreted Features}\label{app:BadFeatures}
\subsubsection{\enquote{references to coffee} (29/30028)}\label{app:BadFeatures_coffee}
\noindent
\textbf{Activation pattern:}\\
\begin{table}[ht!]     
    \small     
    \captionsetup{font=small}     
    \centering     
    \begin{tabular}{p{.96\textwidth}} \hline
        A \redhighlight{red!100}{\strut coff}\redhighlight{red!23.47}{\strut e}eshop is more than just a place to drink—it's a daily ritual for many. Freshly brewed coffee, warm atmosphere, and the right company can turn simple visits into rich, flavorful experiences that feel both comforting and energizing. \redhighlight{red!77.36}{\strut Coff}ins are more than containers; they are symbols of respect and remembrance. \redhighlight{red!90.09}{\strut Coff}ins can be simple and plain or richly decorated, depending on tradition and status. Across cultures, a coffin reflects beliefs about death, and serves as a final gesture of honor for the one who has passed. \\ \hline
    \end{tabular}
    \caption{Activation of feature \enquote{references to coffee} (29/30028) on English text. The feature activates on the token \enquote{coff}. Regardless of the context being either \enquote{coffeeshop} or \enquote{coffin}.}
    \label{tab:activation_coffee}
\end{table}

\noindent
\textbf{Steering Capabilities:} \\
\begin{table}[ht!]   
    \small
    \captionsetup{font=small}
    \centering     
    \begin{tabular}{|p{.12\textwidth}|p{.82\textwidth}|}\hline
        Steering & He told me that, at the funeral, five people lowered the... \\ \hline\hline
        Baseline & ... \textbf{coffin} into the grave but only two could lift it out again.\newline He said he was in a car accident and he had to be cut out of his car with the jaws of life.\newline I asked him how he was doing and he said \enquote{I'm okay, I'm just having a little trouble breathing.} I asked him if he wanted me to call [...] \\
        Coeff $-20$ & ... \textbf{casket} into the ground and then stood there in silence. When he asked them what they were doing, they said they were waiting for him to tell them what to do next.\newline He said that he was embarrassed and didn't know what to say. He was still trying to come to terms with his mother's death. He had never been in charge [...]\\ \hline
    \end{tabular} 
    \caption{Steering capability of feature \enquote{references to coffee} (29/30028). Negatively clamping this feature suppresses the term \enquote{coffin} even though it does not relate to \enquote{coffee} (except for the same beginning of the words).} 
    \label{tab:clamping_20_coffee} 
\end{table}

\newpage
\subsubsection{\enquote{the word \enquote{Morocco} and its variations or related concepts} (18/6893)}\label{app:BadFeatures_Morocco}
\noindent
\textbf{Activation pattern:}\\
\begin{table}[ht!]     
    \small     
    \captionsetup{font=small}     
    \centering     
    \begin{tabular}{p{.96\textwidth}} \hline
        M\redhighlight{red!100}{\strut orr}\redhighlight{red!40.23}{\strut ison} always loved \redhighlight{red!10.2}{\strut mornings} in \redhighlight{red!29.89}{\strut Morocco}, where the \redhighlight{red!7.33}{\strut morning} sun painted the streets with golden light. The \redhighlight{red!56.02}{\strut moral} of his travels, he often thought, was that even the smallest m\redhighlight{red!43.42}{\strut ors}els of experience could teach profound lessons. Over breakfast, he'd savor m\redhighlight{red!47.56}{\strut ors}els of local pastries while reflecting on memories from past adventures, letting each moment shape his \redhighlight{red!32.14}{\strut morals} and outlook. In the quiet of the morning, \redhighlight{red!4.09}{\strut Morrison} felt that life's true richness came from noticing these little m\redhighlight{red!41.54}{\strut ors}els of beauty, connection, and meaning everywhere he went.\\ \hline
    \end{tabular}
    \caption{Activation of feature \enquote{the word \enquote{Morocco} and its variations or related concepts} (18/6893) on English text. The feature activates on \enquote{Morocco}, as well as, the term \enquote{moral} and tokens just entailing \enquote{or}.}
    \label{tab:activation_morocco}
\end{table}

\noindent
\textbf{Steering Capabilities:} \\
\begin{table}[ht!]        
    \small     
    \captionsetup{font=small}     
    \centering          
    \begin{tabular}{|p{.12\textwidth}|p{.82\textwidth}|}\hline
        Steering & A country in North Africa is called... \\ \hline\hline
        Baseline & ... \textbf{Algeria}. It is located on the Mediterranean Sea and is bordered by Tunisia, Libya, Niger, Mali, Mauritania, and Morocco. The country has a population of over 40 million people and is the largest country in Africa. The official language of \textbf{Algeria} is Arabic but French is also widely spoken.\newline \textbf{Algeria} has a diverse geography with mountains, [...] \\
        Coeff $10$ & ... \textbf{Morocco}. It is located on the Mediterranean coast and is bordered by Algeria to the west and the Atlantic Ocean to the south. The capital of \textbf{Morocco} is Rabat, but it also has Casablanca as its largest city. The official language of Morocco is Arabic, but French and Spanish are also spoken there.\newline \textbf{Morocco} has a population of around [...] \\ \hline
    \end{tabular} 
    \caption{Steering capability of feature \enquote{the word \enquote{Morocco} and its variations or related concepts} (18/6893). Positively clamping this feature changes the mentioned North African country from \enquote{Algeria} to \enquote{Morocco}.}
    \label{tab:clamping_10_morocco-b}
\end{table}

\begin{table}[ht!]        
    \small     
    \captionsetup{font=small}     
    \centering          
    \begin{tabular}{|p{.12\textwidth}|p{.82\textwidth}|}\hline
        Steering & My day starts... \\ \hline\hline
        Baseline & ... with a cup of coffee and a quick glance at the news. I'm usually surprised to see how much has happened during the night, and I'm always amazed at how much time people spend on their phones.\newline I love starting my day with a cup of coffee, but sometimes it's hard to find the time to make one. There are so many [...] \\
        Coeff $15$ & ... in the \textbf{mor}ning when I wake up and go to the bathroom and do my \textbf{mor}ning routine. After that, I go to the kitchen and make breakfast. Then I get to school in the \textbf{mor}ning. In the afternoon, I play basketball, play with my friends, or watch TV. In the evening, I go back to my house and then sleep.\newline [...]\\ \hline
    \end{tabular} 
    \caption{Steering capability of feature \enquote{the word \enquote{Morocco} and its variations or related concepts} (18/6893). Positively clamping this feature increases to occurrence of the term \enquote{morning} without a reference to \enquote{Morocco} at any point.} 
    \label{tab:clamping_15_morocco-a} 
\end{table}

\newpage
\subsection{Similarity of Features}\label{app:Similarity}
\subsubsection{Sets of Words for Coffee-Related Features in Layer 18}\label{subsubsec:CategoryTerms}
The following list contains the terms we choose for each of the five coffee-related features in layer 18. The terms were chosen based on the description so that they should generate a high activation for that specific feature:
\begin{itemize}
    \item Feature (18/7546): \enquote{references to drinks, particularly wine and coffee}
    \begin{itemize}
        \item Cabernet, Espresso, Merlot, Flat white, Sauvignon Blanc, Cold brew, Pinot Noir, Latte, Macchiato, Rosé
    \end{itemize}
    \item Feature (18/9463): \enquote{mentions of coffee and related terms}
    \begin{itemize}
        \item Coffee, Caffeine, Brew, Beans, Arabica, Roast, French press, Drip, Grinder, Percolator
    \end{itemize}
    \item Feature (18/15276): \enquote{references to coffee and coffee-related experiences}
    \begin{itemize}
        \item Morning ritual, First sip, Brewing aroma, Wake-up cup, Coffee break, Steaming mug, Late-night study fuel, Café chatter, Bitter comfort, Refueling moment
    \end{itemize}
    \item Feature (18/17350): \enquote{references to cafes and coffee culture}
    \begin{itemize}
        \item Barista, Café ambiance, Third-wave coffee, Latte art, Wi-Fi and workspace, Corner café, Reusable cup, Pour-over station, Coffeehouse playlist, Urban roast bar
    \end{itemize}
    \item Feature (18/28590): \enquote{references to coffee and its cultural significance}
    \begin{itemize}
        \item Coffeehouse intellectualism, Espresso diplomacy, Java as social glue, Coffee as ritual, Daily grind, Global bean trade, Caffeine capitalism, Slow café movement, Coffee as community, Roasting traditions
    \end{itemize}
\end{itemize}

\subsubsection{\enquote{Coffee} Features (Llama 3.1 8B)}\label{app:Features_Coffee}
    \begin{longtable}{p{.06\textwidth} p{.08\textwidth} p{.76\textwidth}}\hline
        \renewcommand{\arraystretch}{1}
        \small      
        Layer & Feature & Description \\ \hline
        0 & 6040 & references to coffee and related beverages \\ 
        0 & 21525 & references to coffee and related products or concepts \\ 
        0 & 25424 & variations of the word "barista" related to coffee making or the coffee industry \\ 
        0 & 26493 &  references to espresso or coffee-related terms \\ 
        1 & 17505 & references to coffee shops and houses \\ 
        1 & 22048 &  words and phrases related to coffee and coffins \\ 
        1 & 31543 & phrases related to beverages, particularly coffee and tea \\ 
        1 & 1068 & references to coffee and related beverages \\ 
        1 & 1912 & topics related to health and wellness, specifically in relation to coffee and its effects \\ 
        2 & 7316 & references to coffee and espresso \\ 
        2 & 10079 &  words related to coffee shops and houses \\ 
        2 & 25767 & mentions of cafes or coffee shops \\ 
        2 & 26873 & terms and phrases related to coffee \\ 
        3 & 777 &  references to coffee and related agricultural products \\ 
        3 & 1936 & references to tables, shops, and houses, particularly in a coffee shop context \\ 
        3 & 2099 & references to Starbucks and coffee-related terms \\ 
        3 & 15768 &  references to coffee-related experiences and businesses \\ 
        3 & 27596 & references to coffee and coffee-related products \\ 
        3 & 29228 &  mentions of cafes and coffee-related establishments \\ 
        4 & 5625 &  mentions of coffee and related beverages \\ 
        4 & 6003 & mentions of cafes or coffee places \\ 
        4 & 1154 &  terms related to coffee and its preparation or consumption \\ 
        5 & 17249 &  mentions of coffee consumption in relation to cancer risk \\ 
        5 & 17365 &  specific terms related to coffee and coffee culture \\ 
        5 & 15894 & references to coffee and related products \\ 
        5 & 32659 & references to Starbucks and related coffee terminology \\ 
        6 & 1006 &  references to the effects of coffee on health, particularly concerning cancer risk \\ 
        6 & 17799 &  references to coffee and tea preparation \\ 
        6 & 17744 &  references to places serving food and beverages, particularly cafes and coffeehouses \\ 
        6 & 16797 &  terms related to coffee and its preparation \\ 
        6 & 15111 &  terms related to coffee and its preparation or consumption \\ 
        6 & 21678 & references to specific entities, particularly numbers and brands related to coffee \\ 
        6 & 25623 & references to coffee \\ 
        7 & 13 &  references to coffee-related terms \\ 
        7 & 14416 & references to coffee and related products or experiences \\ 
        7 & 15261 &  mentions of coffee and related terms \\ 
        7 & 15843 &  references to coffee and its active ingredient, caffeine \\ 
        7 & 23800 & references to social activities that involve beverages, especially coffee and tea \\ 
        7 & 3001 &  words and phrases related to coffee consumption and its effects on health \\ 
        7 & 6348 &  locations and settings related to coffee shops, restaurants, and public spaces \\ 
        7 & 12342 &  terms related to coffee-focused initiatives and their impacts on different communities \\ 
        7 & 30127 &  mentions of tea and coffee \\ 
        8 & 9164 &  terms related to coffee and coffee preparations \\ 
        8 & 2642 &  references to cups of coffee and tea \\ 
        8 & 11299 &  references to cafes and coffee shops \\ 
        8 & 14805 &  references to coffee consumption and its potential health effects \\ 
        8 & 27662 & patterns and relationships in research findings regarding cancer risk and influences of coffee consumption \\ 
        8 & 28820 & terms related to caffeine and coffee consumption \\ 
        8 & 31736 &  references to coffee-related topics and products \\ 
        9 & 22136 & references to coffee consumption and its relationship to health outcomes \\ 
        9 & 26076 &  references to studies examining the correlation between coffee consumption and cancer risk \\ 
        9 & 26466 & references to coffee-related topics and experiences \\ 
        10 & 19221 &  phrases related to the consumption of coffee and its impacts on health \\ 
        10 & 29384 &  references to coffee and coffee-related experiences \\ 
        11 & 19594 & references to coffee consumption and its relationship with cancer risk \\ 
        11 & 23647 &  references to coffee and related products or experiences \\ 
        12 & 5988 & terms related to cancer, coffee consumption, and marijuana usage \\ 
        12 & 8704 & references to coffee and related experiences \\ 
        13 & 9732 & terms related to coffee brewing and its cultural significance \\ 
        13 & 23357 & references to coffee and its variations \\ 
        13 & 11617 & statements related to coffee and its impact on health \\ 
        14 & 3181 &  references to coffee and its related activities or products \\ 
        15 & 6555 &  mentions of coffee consumption and its correlation with health outcomes \\ 
        15 & 3366 &  mentions and discussion of coffee \\ 
        15 & 4922 & references to the effects of coffee on health \\ 
        15 & 17398 &  references to coffee and its preparation \\ 
        15 & 28235 & references to coffee and its related business processes \\ 
        16 & 5015 & references to coffee and coffee-related establishments \\ 
        16 & 23756 &  references to studies and data regarding cancer risk associated with coffee consumption \\ 
        16 & 24458 & references to coffee and its various implications or characteristics \\ 
        16 & 30530 & relationships concerning health studies and associations with coffee consumption \\ 
        16 & 31454 & references to drinking beverages, particularly coffee and tea \\ 
        17 & 8387 & references to coffee and related terms \\ 
        17 & 17320 &  topics related to coffee, including its origins, effects, and cultural significance \\ 
        17 & 30011 &  references to cafés and coffee shops \\ 
        18 & 7546 &  references to drinks, particularly wine and coffee \\ 
        18 & 9463 & mentions of coffee and related terms \\ 
        18 & 15276 &  references to coffee and coffee-related experiences \\ 
        18 & 17350 & references to cafes and coffee culture \\ 
        18 & 28590 &  references to coffee and its cultural significance \\ 
        19 & 12587 & references to coffee \\ 
        19 & 22735 &  references to cafés or coffee shops \\ 
        20 & 2343 &  references to coffee-related experiences and businesses \\ 
        20 & 6891 & references to coffee and its various contexts \\ 
        20 & 30233 &  references to social interactions in coffee shops \\ 
        21 & 28489 &  mentions of coffee and its related contexts \\ 
        21 & 1749 &  references to the coffee industry and related experiences \\ 
        21 & 4914 &  specific brand names or references associated with coffee \\ 
        21 & 24045 & references to coffee and related terms \\ 
        21 & 24657 &  numbers and product details related to coffee and kitchen items \\ 
        21 & 30850 & mentions of cafes and coffee-related establishments \\ 
        22 & 8201 &  references to coffee and its various contexts \\ 
        22 & 10804 & references to coffee production and supply chains \\ 
        22 & 21262 & terms related to coffee and food \\ 
        22 & 21760 &  phrases related to coffee and its experiences \\ 
        23 & 11578 & mentions of coffee and related terms \\ 
        23 & 29983 &  specific terms and phrases related to coffee and barista culture \\ 
        23 & 4333 & variations of the word "coffee." \\ 
        23 & 6885 &  product-related information and details, particularly those associated with coffee grinders and other kitchen appliances \\ 
        23 & 7440 &  terms related to coffee, including references to its preparation and consumption experiences \\ 
        23 & 13942 & mentions of cafes and coffee-related establishments \\ 
        23 & 20675 &  references to coffee and its production \\ 
        23 & 32054 & references to commercial kitchen appliances and coffee machines \\ 
        24 & 7629 &  references to coffee-related products and their cultural significance \\ 
        24 & 21855 &  key features and elements related to a coffee experience and culture \\ 
        24 & 16957 &  terms related to coffee and its preparation or consumption \\ 
        24 & 22301 &  references to coffee products and related terms \\ 
        24 & 28061 & references to cafes and related coffee establishments \\ 
        25 & 14539 &  mentions of coffee-related terms \\ 
        25 & 11817 & references to coffee and café culture \\ 
        25 & 13111 &  references to coffee experiences and related business activities \\ 
        25 & 19316 & information related to the effect of coffee consumption on cancer risk \\ 
        25 & 22482 &  references to coffee culture and its social impact \\ 
        25 & 22974 &  references to cancer and its associated risks, particularly in relation to coffee consumption \\ 
        25 & 27630 &  mentions of coffee and related terminology in various contexts \\ 
        25 & 29299 & references to coffee or related terms \\ 
        26 & 32692 & specific names or brand labels related to coffee \\ 
        26 & 3281 &  mentions of coffee and its variations \\ 
        26 & 10709 &  mentions of coffee \\ 
        26 & 10896 &  references to coffee-making equipment and related terms \\ 
        26 & 18981 & mentions of Starbucks and related coffee culture \\ 
        26 & 19081 & strongly positive sentiments and preferences related to coffee \\ 
        26 & 22463 &  references to coffee-related businesses and their social impact \\ 
        27 & 1729 &  words related to coffee \\ 
        27 & 2746 &  references to coffee-related activities and establishments \\ 
        27 & 3792 &  references to coffee-making equipment and techniques \\ 
        27 & 4442 & sentences that discuss coffee experiences and related cultural practices \\ 
        27 & 11953 &  references to coffee and its cultural significance \\ 
        27 & 29811 &  information related to cancer risk and studies involving coffee consumption \\ 
        28 & 3469 & references to tea and coffee \\ 
        28 & 3737 & references to coffee and coffee-related establishments \\ 
        28 & 18037 &  elements related to the coffee industry and artisanal coffee-making \\ 
        28 & 20185 &  elements of coffee brewing equipment and related terminology \\ 
        28 & 20740 &  references to coffee culture and experiences \\ 
        28 & 24964 & references to beverages, particularly those involving alcohol and coffee \\ 
        28 & 32127 &  terms related to coffee and its production, particularly focusing on specific origins and qualities \\ 
        29 & 877 &  economic or statistical information related to coffee production and trade \\ 
        29 & 1349 & references to coffee and coffee-related establishments \\ 
        29 & 6546 &  references to coffee and its cultural significance \\ 
        29 & 6762 &  mentions of tea and coffee \\ 
        29 & 10040 &  expressions related to the coffee experience and community engagement \\ 
        29 & 23015 &  references to cancer and its relationship with coffee consumption \\ 
        29 & 29263 & information related to coffee brewing methods \\ 
        29 & 30028 & references to coffee \\ 
        30 & 13620 &  references to coffee-related terminology and equipment \\ 
        30 & 14209 & references to beverages, particularly coffee and tea \\ 
        30 & 18726 & references to coffee and related establishments \\ 
        30 & 23901 &  references to coffee and its production or distribution \\ 
        30 & 24229 & terms related to the coffee industry and its business practices \\ 
        30 & 27565 &  references to coffee-making equipment and techniques \\ 
        30 & 31724 & references to coffee and related terms \\ 
        31 & 4216 & references to coffee, particularly its origins and impacts related to trade and economics \\ 
        31 & 5840 &  references to methods of brewing coffee \\ 
        31 & 11770 &  references to coffee and related experiences \\ 
        31 & 14507 &  references to various coffee-related products and services \\ 
        31 & 22738 & references to coffee and coffee-related establishments \\ 
        31 & 26080 & references to cups of beverages, especially tea and coffee \\ \hline
        \captionsetup{font=small}
        \caption{All Features of Llama 3.1 8B (Llama Scope for residual stream) containing the word \enquote{coffee}.}
    \end{longtable}

\subsubsection{Activation Patterns of \enquote{references to coffee} Features}\label{app:Activation_referencesToCoffee}
\begin{table}[ht!]
    \small
    \captionsetup{font=small}
    \centering
    \begin{tabular}{p{.96\textwidth}} \hline
         As the barista carefully grinds the aromatic \redhighlight{red!20.54}{\strut espresso} \redhighlight{red!5.79}{\strut beans} and tamps them into the portafilter before pulling a silky shot of ristretto, the bustling cafe fills with the comforting scent of roasted Arab\redhighlight{red!19.55}{\strut ica}, while patrons sip their cappucc\redhighlight{red!12.19}{\strut inos} from ceramic mugs, chatting about fair-trade plantations, latte art, crema, and the timeless ritual that makes \redhighlight{red!44.06}{\strut coffee} not just a beverage but a global symbol of morning routines, social connection, and mindful indulgence.\\ \hline
    \end{tabular}
    \caption{Activation of feature \enquote{references to coffee} (6/25623) on English text. The feature related to coffee activates on and around coffee terms. The closer the words are to coffee the higher the activation. Higher opacity indicates stronger activation.}\label{tab:activation_6-25623_english}
\end{table}
\begin{table}[ht!]
    \small
    \captionsetup{font=small}
    \centering
    \begin{tabular}{p{.96\textwidth}} \hline
         As the barista carefully grinds the aromatic \redhighlight{red!56.44}{\strut espresso} beans and tamps them into the portafilter before pulling a silky shot of ristretto, the bustling cafe fills with the comforting scent of roasted Arabica, while patrons sip their capp\redhighlight{red!20.17}{\strut ucc}\redhighlight{red!24.26}{\strut inos} from ceramic mugs, chatting about fair-trade plantations, lat\redhighlight{red!35.89}{\strut te} art, crema, and the timeless ritual that makes \redhighlight{red!100}{\strut coffee} not just a beverage but a global symbol of morning routines, social connection, and mindful indulgence.\\ \hline
    \end{tabular}
    \caption{Activation of feature \enquote{references to coffee} (19/12587) on English text. The feature related to coffee activates on and around coffee terms. The closer the words are to coffee the higher the activation. Higher opacity indicates stronger activation.}\label{tab:activation_19-12587_english}
\end{table}
\begin{table}[ht!]
    \small
    \captionsetup{font=small}
    \centering
    \begin{tabular}{p{.96\textwidth}} \hline
         As the barista carefully grinds the aromatic espresso beans and tamps them into the portafilter before pulling a silky shot of ristretto, the bustling cafe fills with the comforting scent of roasted Arabica, while patrons sip their cappuccinos from ceramic mugs, chatting about fair-trade plantations, latte art, crema, and the timeless ritual that makes coffee not just a beverage but a global symbol of morning routines, social connection, and mindful indulgence.\\ \hline
    \end{tabular}
    \caption{Activation of feature \enquote{references to coffee} (29/30028) on English text. The feature's descritpion is related to coffee. The feature, however, does not activate on and around coffee terms. In Section \ref{subsubsec:BadFeatures}, we show that it activates on the \enquote{coff} token instead (note that \enquote{coffee} is its own token).}\label{tab:activation_29-30028_english}
\end{table}

\newpage
\subsubsection{Steering of Coffee-Related Features}\label{app:Steering_Coffee_all}
    \begin{longtable}{|p{.1\textwidth} p{.11\textwidth} | p{.72\textwidth}|} \hline
        \renewcommand{\arraystretch}{1.5}
        \small
        Steering & Feature & My favorite drink is... \\ \hline\hline
        Baseline & ~ & ... a margarita but my favorite cocktail is a mojito. I have a lot of memories of enjoying mojitos with my friends and family. \newline
        I love to cook and bake, but I am not very good at it. My best dish is probably pasta carbonara. I make it every time I go to Italy. \newline
        My favorite thing to \\ \hline
        Coeff $1$ & 6/25623 & ... a margarita but my favorite cocktail is a mojito. I have a lot of different recipes for mojitos, but this one is my favorite. It's simple and packs a punch.  \newline The name of this cocktail, ``The mojito'', comes from the Spanish word ``mocas'' which \\ 
        ~ & 19/12587 & ... a margarita but my favorite cocktail is a mojito. I have a lot of memories of enjoying mojitos with my friends and family.  \newline I love to cook and bake, but I am not very good at it. My best dish is probably pasta carbonara. I make it every time I go to \\ 
        ~ & 29/30028 & ... a margarita but my favorite cocktail is a mojito. I have a lot of different recipes for mojitos, but this one is my favorite. It's simple and packs a punch.  \newline The name of this drink, ``Mojito,'' means ``little drink'' in Spanish. It's made with \\ \hline
        Coeff $2$ & 6/25623 & ... a margarita but my favorite cocktail is a mojito. I have a lot of different coffee drinks in the morning, but I always have it with milk. I don't really like tea.  \newline I like to cook and bake. My favorite food is pasta and pizza. I love to eat meat, fish, \\ 
        ~ & 19/12587 & ... a margarita but my favorite cocktail is a mojito. I have a lot of memories of enjoying mojitos with my friends and family.  \newline I love to cook and bake, but I am not very good at it. My best dish is probably pasta carbonara. I make it every time I go to \\ 
        ~ & 29/30028 & ... a margarita but my favorite cocktail is a mojito. I have a lot of different recipes for mojitos, but this one is my favorite. It's simple and packs a punch.  \newline The name of this drink, ``Mojito,'' means ``little drink'' in Spanish. It's made with \\ \hline
        Coeff $5$ & 6/25623 & ... a c late but and I love, and is my favorite. Coffee, I love coffee and, coffee is my coffee. Our, we, coffee.  \newline Coffee Coffee Coffee  \newline Coffee, coffee coffee.  \newline Coffee is delicious  \newline coffee, tea, coffee  \newline coffee coffee,  \newline coffee coffee!  \newline Coffee is the best  \newline Coffee Coffee \\ 
        ~ & 19/12587 & ... coffee. I usually drink it black and without cream. I like to get my caffeine fix in the morning before I start working. This helps me stay awake and alert throughout the day.  \newline I love coffee because it is delicious and energizing. It also has many health benefits, such as reducing the risk of heart disease and \\ 
        ~ & 29/30028 & ... a margarita but my favorite cocktail is a mojito. I have a lot of different recipes for mojitos, but this one is my favorite. It's simple and packs a punch.  \newline The name of this drink comes from the mojito plant that grows in Puerto Rico. The plant has small white \\ \hline
        Coeff $10$ & 6/25623 & ... in flavor beans,, and, and, beans.  \newline  beans coffee coffee Coffee Coffee Coffee coffee coffee cup coffee coffee Coffee Coffee coffee coffee.  \newline , coffee. tea coffee cafe Coff Coffee  \newline . Cafe Coffee C.  \newline  Coffee black  \newline  Espresso C Coffee tea Breakfast coffee coffee Breakfast  \newline  drink  \newline  breakfast caffeine espresso c breakfast brew tea Tea cup \\ 
        ~ & 19/12587 & ... coffee. I usually drink it black and without cream. I like to get my caffeine fix in the morning before I start working. This helps me stay alert and focused throughout the day.  \newline I love coffee because it is a great way to wake up in the morning. It has caffeine, which is a stimulant that helps \\ 
        ~ & 29/30028 & ... coffee. I usually drink my coffee black, but I like to add creamer to my coffee when I'm in the mood for something sweet. I've tried a few different brands of creamers, and I've found that some are better than others. In this article, I'll share with you my top 5 \\ \hline
        Coeff $20$ & 6/25623 & ... and,,,, and, and, break.  \newline  beans coffee coffee Coffee, grounds coffee coffee break coffee  ... beans coffee coffee.  \newline , we black morning cream before with and in first before.  \newline  coffee.  \newline  healthy shop tables with ..., cups early coffee coffee also black sweet first breakfast on after. breakfast tables tea black coffee \\ 
        ~ & 19/12587 & ... a simple one: a cup of black, with a bit of cream. I love to add some grounds and cream, and then just sit and drink it. It's the perfect way to wake up in the morning. I also like to make it at home, because it's so easy to do.  \newline Making coffee at \\ 
        ~ & 29/30028 & ... coffee. I love the smell, taste, and feel of coffee. Coffee has a lot of benefits for our health like it can increase our energy and it can help to boost our brain function. I love to drink coffee with my friends or family in the morning or afternoon.  \newline Coffee is one of the most popular drinks in \\ \hline
        Coeff $30$ & 6/25623 & ...,,,,,-,-co-b-b-c-co-co-c-dr-c grounds-co-dr break coffee coffeeacial-com-col-gl-co-c-gl \&-menu-b coffee-client coffeeCoffeeoffeeerrar474 joeï¿½ Joeorders ï¿½ ï¿½474bet-com Arab coffee joecoffee punch-script-gl-office ï¿½ cafÃ©\_bit-go\_br groundsacialå†² \\ 
        ~ & 19/12587 & ... and... \newline Shop shop shops and shop shops, and then shop  \newline houses, and grounds, and break shops...  \newline and beans, and beans, and beans, beans--- bean time  \newline and beans break break break break --break time. â€”house  \newline and grounds, and beans, beansâ€”beans--- bean tables-table table tables \\ 
        ~ & 29/30028 & ... coffee. I love the smell, taste, and feel of coffee. Coffee is a big part of my life and has been for many years. I have coffees in the morning, coffees in the afternoon, coffees at night.  \newline Coffee is one of those things that can be enjoyed at any time of the day \\ \hline
        Coeff $50$ & 6/25623 & \raisebox{-0.85\totalheight}{\includegraphics[width=1\linewidth]{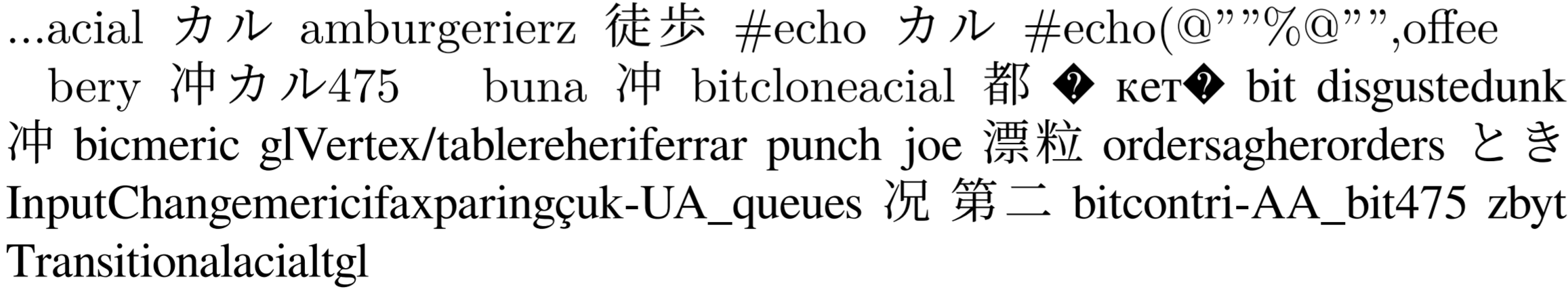}} \\
        ~ & 19/12587 & ... and beans beans beans bean grounds beans grounds bean grounds beans beanshousehouseshousehouse grounds beans coffee break break break break bean grounds coffee grounds beans coffee break shopScriptScript shops grounds shops shops coffee joe joe bean Joe mug cups mug cuphousehouse-table table-table-te-table-table tables-table-table filter filters makers tables tea/es/es \\ 
        ~ & 29/30028 & ... coffee. I love coffeeseinsinsinsinsinsinsinsin coffeeeseyeeseyeeseyeeseyeeseyeeseyeeseyeeseyeeseyeeseyeeseyeeseyeeseyeesyeh Coffe Coffe Coffe Coffe Coffe Coffe Coffe Coffe Coffe \\ \hline
        \captionsetup{font=small}
        \caption{Output for selected positive steering coefficients for all features of Llama 3.1 with the description \enquote{references to coffee}. Although having the same description all three features exhibit different steering results.  [Note that some outputs are displayed as pictures to ensure the correct rendering of uncommon characters.]}\label{tab:clamping_referencesCoffee}
    \end{longtable}

    \begin{longtable}{|p{.11\textwidth} p{.1\textwidth} | p{.72\textwidth}|} \hline
        \renewcommand{\arraystretch}{1.5}
        \small
        Steering & Feature & My favorite drink is... \\ \hline\hline
        Baseline & ~ & ... a margarita but my favorite cocktail is a mojito. I have a lot of memories of enjoying mojitos with my friends and family. \newline
        I love to cook and bake, but I am not very good at it. My best dish is probably pasta carbonara. I make it every time I go to Italy. \newline
        My favorite thing to \\ \hline
        Coeff $1$ & 7546 & ... a margarita but my favorite cocktail is a mojito. I have a margarita almost every day during the summer and I have a mojito when it's cold out. I love the fresh mint and lime in a mojito. It's so refreshing on a hot summer day. \newline
        I don't know what it is about \\ 
        ~ & 9463 & ... a margarita but my favorite cocktail is a mojito. I have a lot of memories of enjoying mojitos with my friends and family. \newline
        I love to cook and bake, but I am not very good at it. My best dish is probably pasta carbonara. I make it every time I go to Italy. \newline
        My favorite thing to \\ 
        ~ & 15276 & ... a margarita but my favorite cocktail is a mojito. I have a lot of different recipes for mojitos, but this one is my favorite. It's simple and packs a punch. \newline
        The name of this drink, “Mojito,” means “little water” in Spanish. This cocktail has been around since the 16th \\ 
        ~ & 17350 & ... a margarita but my favorite cocktail is a mojito. I have a lot of memories of sitting in the backyard with my mom and drinking mojitos. It's a really simple drink to make and it's refreshing, so it's perfect for summer. I've been making them at home for years, and they're always a hit \\ 
        ~ & 28590 & ... a margarita but my favorite cocktail is a mojito. I have a lot of different recipes for mojitos, but this one is my favorite. It's simple and packs a punch. \newline
        The name of this drink comes from the mojito plant that grows in Puerto Rico. The plant has small white flowers that are used in many \\ \hline
        Coeff $2$ & 7546 & ... a margarita but my favorite cocktail is a mojito. I'm not sure why but I just love the flavor of mint and lime together. \newline
        I'm pretty sure that if you were to ask me what ..., I'd say margaritas. But I'm also pretty sure that if you were to ask me what my \\ 
        ~ & 9463 & ... coffee. I usually drink my coffee black, but sometimes I like to add a little milk or cream to it. I also like to add sugar or sweetener to my coffee. \newline
        Coffee is a brewed beverage with a distinct aroma and flavor. It is one of the most popular drinks in the world, and it has been around for centuries. Coffee beans \\ 
        ~ & 15276 & ... coffee. I usually drink my coffee black, but sometimes I like to add a little milk or cream to it. I also like to add sugar or sweetener to my coffee. \newline
        Coffee is a brewed beverage with a distinct flavor. It is usually made from roasted coffee beans, which are the seeds of the Coffea plant. The beans are ground and \\ 
        ~ & 17350 & ... coffee. I usually drink my coffee black, but sometimes I like to add a little milk or cream. If you're looking for a delicious and healthy way to enjoy your coffee, consider adding some milk. \newline
        There are many different types of milk that you can use in your coffee, and each one has its own unique flavor and benefits. In this blog \\ 
        ~ & 28590 & ... a margarita but my favorite cocktail is a mojito. I have a lot of different recipes for mojitos, but this one is my favorite. It's simple and packs a punch. \newline
        The name of this cocktail, “The mojito”, comes from the Spanish word “mocas” which means “to mix”. The \\ \hline
        Coeff $5$ & 7546 & ... a glass of cold water and I usually drink it with my meals. I have a lot of different types of water, but my favorite is mineral water. \newline
        I'm not sure if this is the best way to describe it, but mineral water is the most common type of water in the world. It's also the most expensive. There are many different \\ 
        ~ & 9463 & ... coffee. I usually drink it black and without cream. I like to get my caffeine fix in the morning and in the afternoon. This is why I am always on the lookout for a good cup of coffee. \newline
        I love coffee. It's one of my favorite drinks. I like to make it at home, but I also enjoy going out for a \\ 
        ~ & 15276 & ... coffee. I usually drink it black, but I also like it with cream or milk. I love the way the aroma of freshly brewed coffee fills my kitchen in the morning and the first sip of coffee in the morning is one of my favorite things. It's a little bit like waking up to a warm hug. \newline
        I have been drinking coffee for many \\ 
        ~ & 17350 & ... coffee. I usually drink my coffee black, but sometimes I like to add a little milk or cream. If you're looking for a delicious and healthy way to enjoy your coffee, consider adding some milk. \newline
        There are many different types of milk that you can use in your coffee, and each one has its own unique flavor. Some people prefer almond milk \\ 
        ~ & 28590 & ... coffee. I usually drink it black and without sugar. I like to get my caffeine fix in the morning, but during the day, I have a few cups of tea instead. \newline 
        I love all kinds of tea, but my favorites are green and white teas. I also enjoy rooibos tea, which is an herbal tea that comes from South \\ \hline
        Coeff $10$ & 7546 & ... a glass of wine with my dinner and a beer with my friends. I have a lot of wine, but I like it with my friends. I'm sure it's good for me, but I don't like it with my friends. \newline
        I'm not the only one who likes to drink. My friend, the bartender at the bar, is always \\ 
        ~ & 9463 & ... coffee. I drink it every morning and sometimes in the afternoon. I'm not a fan of flavored coffee, but I do like to add cream and sugar to my cup. My favorite coffee brand is Starbucks, but I also like to try other brands as well. \newline
        I'm not a big fan of tea, but I do enjoy it sometimes. My \\ 
        ~ & 15276 & ... coffee. I drink it every day and I love to brew it at home. I have a variety of coffee makers, but my favorite is the French press. \newline
        The French press is a simple way to make a great cup of coffee. It's easy to use, and it produces a rich, flavorful brew. \newline
        The coffee grounds are steeped in hot \\ 
        ~ & 17350 & ... coffee. I usually drink my coffee black, but sometimes I like to get a little crazy and add some flavor to it. One of my favorite ways to do this is with a chai latte. This is a delicious and warming drink that is perfect for the cold weather. \newline
        This chai latte recipe is simple and easy to make. All you need \\ 
        ~ & 28590 & ... a ""Duck Pong"" that is a dark roast with milk and honey. It is very smooth and rich. \newline
        I like to drink the ""Duck Pong"" in the evening when I am relaxing in my garden. The tea has a very smooth taste and it goes well with the honey and milk. \newline
        The tea leaves are from the Y \\ \hline
        Coeff $20$ & 7546 & ... a glass of wine from my wine and water while I am reading. I have a lot of wine and water in the morning while I am celebrating. I have a few with my friends and some with the red wine. The red wine is good for your white tea and it is on the go at night. \newline
        The coffee is good for your whiskey, \\ 
        ~ & 9463 & ... and ice, and sometimes I can make it in a shophouse. But, I need to break break for making beans. So, this maker machine is really good for me. \newline
        It was very good for me because it was easy to make and it was also really sweet. It was a great cup and I made it at the shop house. \newline
        This \\ 
        ~ & 15276 & ... coffee. I drink it every morning and I usually brew it using a machine. I have a coffee machine that is made by a brand called coffee by flavor. \newline
        Coffee by flavor is a brand that has been around for more than 10 years, and it's one of the best on the market. It makes a great cup of coffee using only two \\ 
        ~ & 17350 & ... coffee. I usually drink it to get a quick boost in the morning or later in the afternoon when I need to get work done and can't afford to sit down for a full meal. I've been working on getting my laptop, reading, and coffee spot at my local cafe. The place is called ""The Think"" and it's a great \\ 
        ~ & 28590 & ... the ""C.T.G.E. or the C.T.G.E. B.E.I. (Black Estate 1997) from the Black Estate of Yr. T.G.B.E.S. (Taste Green Yellow Broken Leaf S.F.G.E. (S.F.G.E. is a tea grade tea leaf from the H.B.P.C.A \\ \hline
        Coeff $30$ & 7546 & ... a "" while wine from my wine and s water with my beer. I have a to while and water with my wine and while. \newline
        I have a wine with my to and \newline
        and water with my to \newline
        and water with s \newline
        with s mo \newline
        with mo sweet \newline
        on s red or \newline
        red tea or \newline
        tea black or \newline
        black wine \\ 
        ~ & 9463 & ... and... \newline
        / shop talk / beans gr gr gr beans \newline
        / ro grounds mug grounds beans beans break break grinder grinder bean grounds beans bean, grounds, shop, machine, and, shops, andhouse, k shop)househousehousehouse coffee coffee coffee bean coffee beans beans ro ro filter grinding machines tables tea table makers machines machine beans-be grounds-co \\ 
        ~ & 15276 & ... coffee flavor coffee coffee coffee and I want to drink brew coffee flavor coffee brew coffee brew coffee brewing cup water brew water french americon brewed brewing best brewing water brewing coffee grind grind \newline
        Coffee, how good it is, how delicious it is, and the most brewed single drink per per brewed filter americon brew tea leaf tea cup per brewed using grounds gr \\ 
        ~ & 17350 & ... a ""coffee with"" and I usually order a coffee with or to go, or to sit and have a ""coffee with"" while reading or reading while drinking. \newline
        For a few people, the coffee is the place to sit and read. It's a place to sit and read. The place to sit and read, and the people are sitting. \newline
         \\ 
        ~ & 28590 & ... the "" C. E. T. Tea B. E . W . S. T. F . G . S. P... 1 Tst Estate Tea Estate Estate Tea Auctioner Grade Estate orthodox estate orthodox pek teas leaf orthodox tea leaf tea orthodox Leaf tea Organic Buk estate estate pekica tea pek tea leafer grade grade pekica cupr \\ \hline
        Coeff $50$ & 7546 & \raisebox{-0.85\totalheight}{\includegraphics[width=1\linewidth]{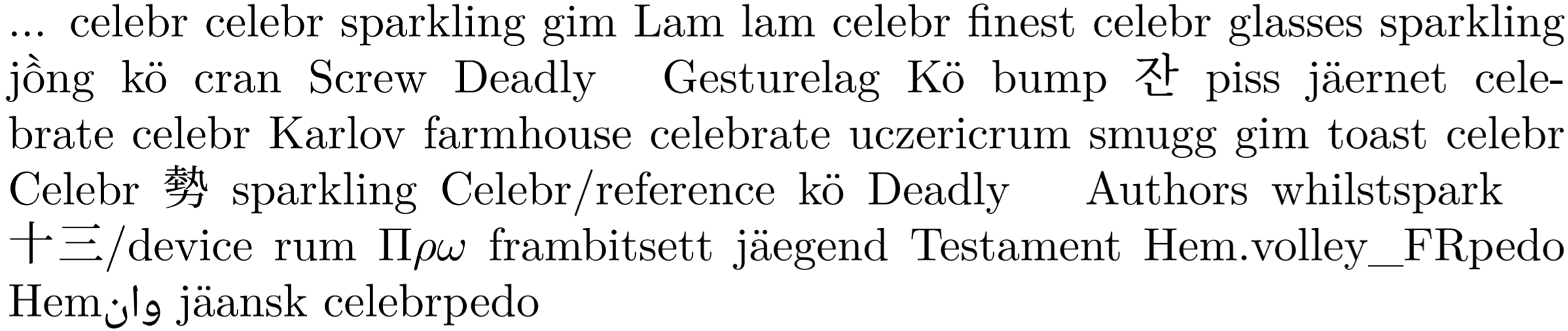}}\\
        ~ & 9463 & \raisebox{-0.85\totalheight}{\includegraphics[width=1\linewidth]{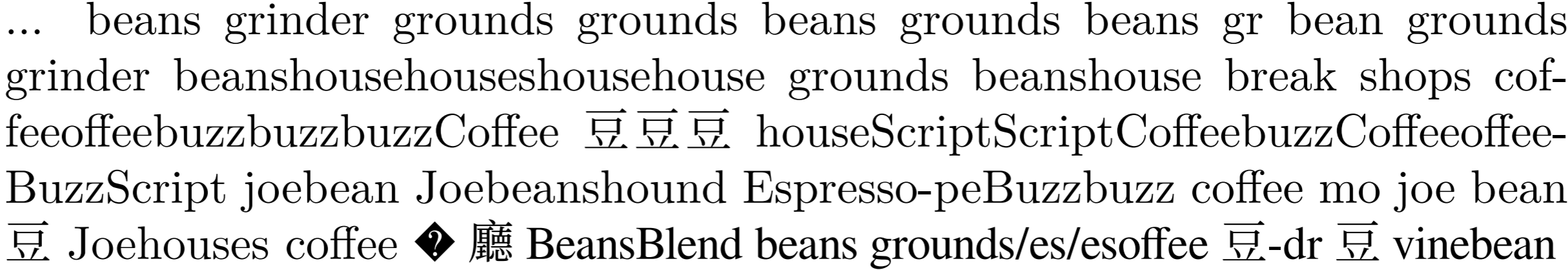}} \\
        ~ & 15276 & \raisebox{-0.85\totalheight}{\includegraphics[width=1\linewidth]{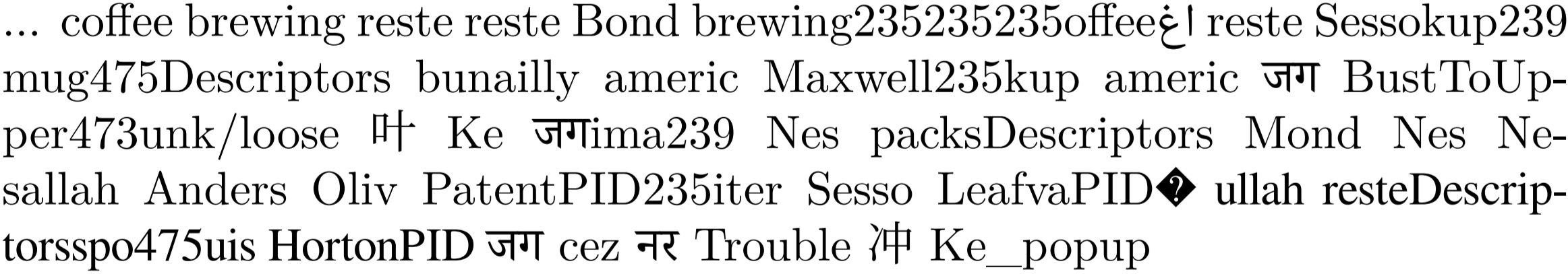}} \\
        ~ & 17350 & \raisebox{-0.85\totalheight}{\includegraphics[width=1\linewidth]{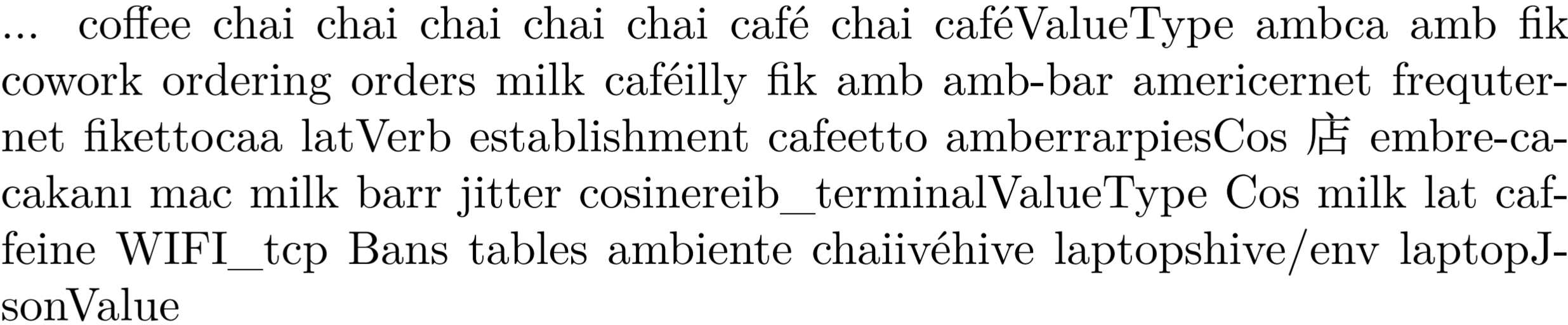}} \\
        ~ & 28590 & \raisebox{-0.85\totalheight}{\includegraphics[width=1\linewidth]{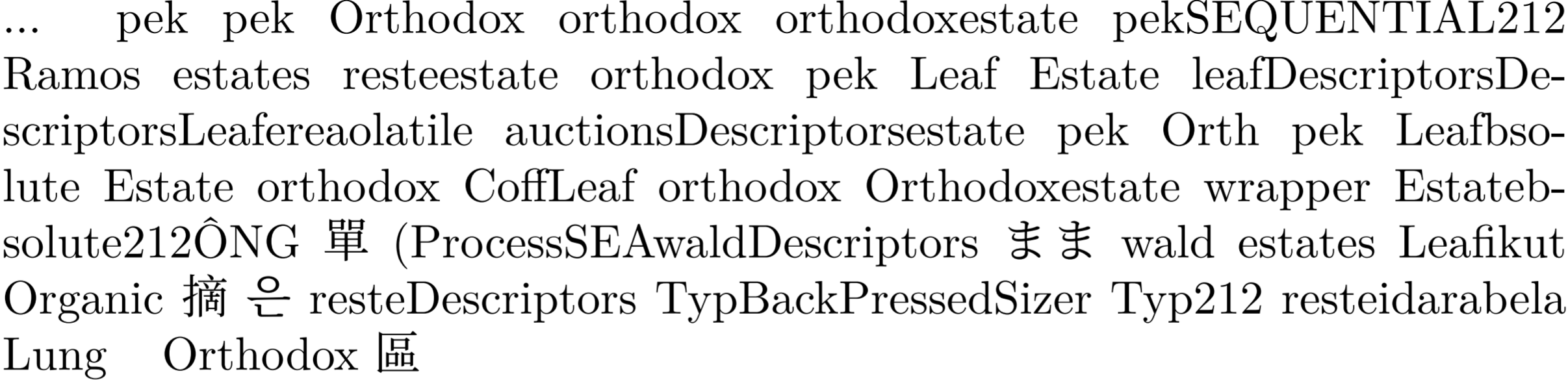}} \\ \hline
        \captionsetup{font=small}
        \caption{Selected positive steering strengths of all features containing the word \enquote{coffee} in layer 18. [Note that some outputs are displayed as pictures to ensure the correct rendering of uncommon characters.]}\label{tab:clamping_coffee_layer18_positive}
    \end{longtable}

\newpage
\begin{longtable}{|p{.12\textwidth} p{.1\textwidth} | p{.72\textwidth}|} \hline
        \renewcommand{\arraystretch}{1.5}
        \small
        Steering & Feature & My ideal breakfast consists of... \\ \hline\hline
        Baseline & ~ & ... a large serving of fruit and a cup of coffee. I love fruit for breakfast because it is so easy to prepare and it gives me the energy that I need to start my day. My favorite fruits are strawberries, blueberries, raspberries, and blackberries. I also like to eat mangoes, bananas, oranges, and apples. \newline
        I always \\ \hline
        Coeff $-1$ & 7546 & ... a large serving of scrambled eggs, an English muffin, and a cup of coffee. I know this is not the best way to start the day, but it is what my body craves. And I am not alone. The average American consumes 35\% of their daily calories at breakfast. This means that if you consume 2,000 \\ 
        ~ & 9463 & ... a large serving of fruit and vegetables, a protein source like eggs, nuts or seeds, a healthy fat source such as avocado or nut butter and a carb source like quinoa or toast. I tend to mix up my breakfast every day so that I don't get bored. I also make sure to have something to drink with my breakfast, usually water \\ 
        ~ & 15276 & ... a large serving of fruit and a bowl of oatmeal. I love oatmeal, and I love the way it makes me feel. It's a great way to start the day, and it's also healthy. I usually make my oatmeal with almond milk, but sometimes I'll make it with coconut milk or even soy milk. It's always \\ 
        ~ & 17350 & ... a large serving of fruit and vegetables, a protein source like eggs, and some sort of whole grain. Â I try to avoid refined carbohydrates because they make me feel sluggish and they don't keep me full for long. Â I also try to avoid processed foods because I'm trying to eat as clean as possible. \newline
        My favorite breakfast is a \\ 
        ~ & 28590 & ... a large serving of scrambled eggs, a piece of bacon, and some fresh fruit. I like to use organic eggs and try to find the best quality bacon I can. Bacon is one of those foods that you can find in every grocery store, but most people don't know how to make it at home. This recipe will show you how to make \\ \hline
        Coeff $-2$ & 7546 & ... a large serving of scrambled eggs, an English muffin or two, and a side of bacon. I'll even add some cheese to the eggs if I'm feeling particularly greedy. \newline
        But, if I was forced to choose one thing for breakfast, it would have to be bacon. I love the smokiness and the salty flavor that bacon brings to \\ 
        ~ & 9463 & ... a large serving of fruit and vegetables, a protein source like eggs, and some healthy fats. I love to eat a big bowl of freshly cut fruit with some eggs or yogurt in the morning. I also like to have some avocado toast if I'm feeling extra hungry! For lunch I usually go for something light like a salad or soup, but if \\ 
        ~ & 15276 & ... a large serving of fruit and vegetables, a protein source like eggs, and some healthy fats. I love to eat a big bowl of mixed berries with sliced avocado and a fried egg on top. This is the perfect combination for me, but you can choose the fruits and vegetables that work best for you. \newline
        This is an easy recipe that can be made \\ 
        ~ & 17350 & ... a large serving of fruit and vegetables, a protein source like eggs, chicken or fish, a healthy fat source such as avocado or coconut oil and a carb source like quinoa or sweet potato. I love to make up my own smoothies with fresh fruit and vegetables. I also love to make breakfast bowls using the ingredients above. There are so many \\ 
        ~ & 28590 & ... a large serving of scrambled eggs, an English muffin with avocado and smoked salmon, and a cup of coffee. I try to avoid the carbs as much as possible, and when I do have them it is usually in the form of a fruit smoothie or some oatmeal. I love to cook and bake, so breakfast is my favorite meal. \newline
         \\ \hline
        Coeff $-5$ & 7546 & ... a large serving of scrambled eggs, topped with a generous amount of chopped fresh herbs, layered on top of a bed of crispy potatoes and smothered in a generous amount of cheese. I love the way the eggs cling to the potatoes and the way the cheese gets all melty on top. It's such an easy breakfast recipe that can be made \\ 
        ~ & 9463 & ... a large serving of fruit and vegetables, a protein source like eggs, and some healthy fats. I love to eat my veggies in the form of a frittata or omelet, but I also like to make sure I get enough fruit in my diet. Today I made this delicious mango breakfast bowl which combines both fruit and vegetables with some healthy \\ 
        ~ & 15276 & ... a large serving of fruit and vegetables, a protein source like eggs, chicken or fish, a healthy fat source such as avocado or nut butter and a carb source like sweet potato or toast. I tend to mix up what I eat every day so it doesn't get boring. \newline
        Sometimes I make this breakfast bowl which is super filling and takes no time at \\ 
        ~ & 17350 & ... a large serving of fruit and vegetables, a protein source like eggs, chicken or fish, a healthy fat source such as avocado or olive oil and a carb source like quinoa or sweet potato. I love to make up my own combinations of foods to create the perfect breakfast for my body. Here are some of my favorite breakfast recipes that are quick and \\ 
        ~ & 28590 & ... a large serving of eggs and toast, with a side of bacon. I'm not sure why I love eggs so much, but I do. I would eat eggs for every meal if I could. And if you're like me, you'll want to have some egg recipes in your arsenal. If you've never tried making an omelet before \\ \hline
        Coeff $-10$ & 7546 & ... a few slices of bread, some butter, and a handful of different toppings. I love to get creative with my toppings, and I've found that there are so many different things that you can do to make your bread topping even better. \newline
        You can use anything from sweet to savory ingredients, or you can even mix them together. There are also tons \\ 
        ~ & 9463 & ... a large serving of fruits and vegetables, a protein, and a little starch. I eat a variety of grains such as quinoa, farina, whole grain oats, and whole wheat bread. I love to make my own granola so that I can control the amount of sugar in it. My favorite breakfast is an open-faced sandwich using sliced avocado \\ 
        ~ & 15276 & ... a large serving of fruit and vegetables, a healthy helping of protein, and a generous portion of fat. Â I try to avoid the breads, pastries, and cereals that are so prevalent in the normal diet. Â I am currently working on my first book on the topic of health and fitness, and I've been trying to \\ 
        ~ & 17350 & ... a large serving of fruits and vegetables, a protein-rich source and some healthy fats. I like to use organic produce as much as possible, but that is not always an option when it comes to the cost. I have found that frozen fruits and vegetables are just as nutritious as fresh produce, so I use them whenever I can. When using frozen foods \\ 
        ~ & 28590 & ... a large serving of eggs and toast, with a side of pancakes. I'm not sure why I never thought to combine the two, but I did it last night and it was a huge success. And guess what? You can make them in the same pan! It's like having two meals at once. \newline
        The only problem with this recipe is that \\ \hline
        Coeff $-20$ & 7546 & ... a set of bags that includes a bag for the books, and another one for the clothes. The bags are supposed to be made from materials that are easy to clean, and they must be able to hold all the items needed for the project. They also have to be able to store all the required materials, like bags, shoes, clothes, and \\ 
        ~ & 9463 & ... a large variety of fruits and vegetables that can be easily found in my local grocery. I enjoy adding some natural protein to my daily diet, but I prefer to include it in my subsequent meals. I am not a strict adherer of any nutritionist's guidelines, nor do I require the inclusion of any form of nutritional information. My approach is to \\ 
        ~ & 15276 & ... a large plate of food and I want to be able to shuffle around all of them to make sure I have the right combination of things that I can eat. So my new idea for the next version of the All-Wrap is to have a large number of sections that can be shuffled around. The problem is that I have no idea how I could \\ 
        ~ & 17350 & ... a large number of dishes and I want to do everything myself. For example, I have a garden and I can grow my own vegetables, I have a huge farm to raise my own animals and I have the best fishing pond. The only thing left is the river to catch my own fish. If you want to see what I'm talking about, \\ 
        ~ & 28590 & ... a meat-and-sauce-entree, a piece of junky crap (e.g., a frozen one-and-a-half-horse-shit-with-a-sprout-on-top), and some kind of no-nutrition garbage like a stupid-ass-ass-ass-ass-ass-ass-ass-ass-ass-ass-ass-ass-ASS-ASS-ASS-ASS- \\ \hline
        Coeff $-30$ & 7546 & ... a set of bags that includes 5, 6 or even 7. The main 5 . 6 . 7 .1 .2 .3 .4 .5 .6 .7 .1 .2 .3ascalaidas , bagasasas , bagasas , bagsasas , bagsasas , bagasas \\ 
        ~ & 9463 & ... a pool of s \newline
        neathne s \newline
        nest \newline
        nnecennlene \newline
        nlerrle \newline
        nrennegcenst- \newline
        ndsmr \newline
        n$\backslash$ \newline
        renenantgtscendn$\backslash$ \newline
        etsnencencnormcnexclestwert$\backslash$ \newline
        Cwrlenrstonlltemenantancementmcuncctipse \\ 
        ~ & 15276 & ... a large enough size of the old and the new with all the old and new in a very good combination. The new is made by the new and old are made by all in a very good combination. \newline
        The name of the new is not to be called as one of the today's, or any other, and I have not been able to find \\ 
        ~ & 17350 & ... some sort of number of number I have for my personal . I do not have to do the . I must for the . The has to be that. I will for and . The should be ARIKRARIKRARIKRARIKRAI-.M................A................M........IAMMMMMMAMMMA-\newline MMMMAM \\ 
        ~ & 28590 & ... a non-nurel-need-need-a-neo-need-a-neo-ne-ne-nurel-need-a-nurel-need-a-nurel-ne-ne-ne-nurel-ne-ne-ne-nure-a-nur-e-nur-e-nur-et-et-et-et-et-et-et-et-et-rere-rere \\ \hline
        Coeff $-50$ & 7546 & \raisebox{-0.85\totalheight}{\includegraphics[width=1\linewidth]{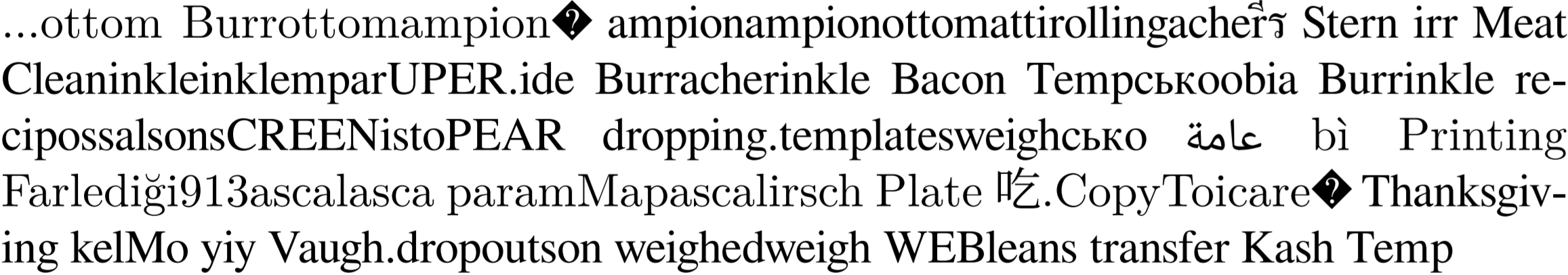}} \\
        ~ & 9463 & \raisebox{-0.85\totalheight}{\includegraphics[width=1\linewidth]{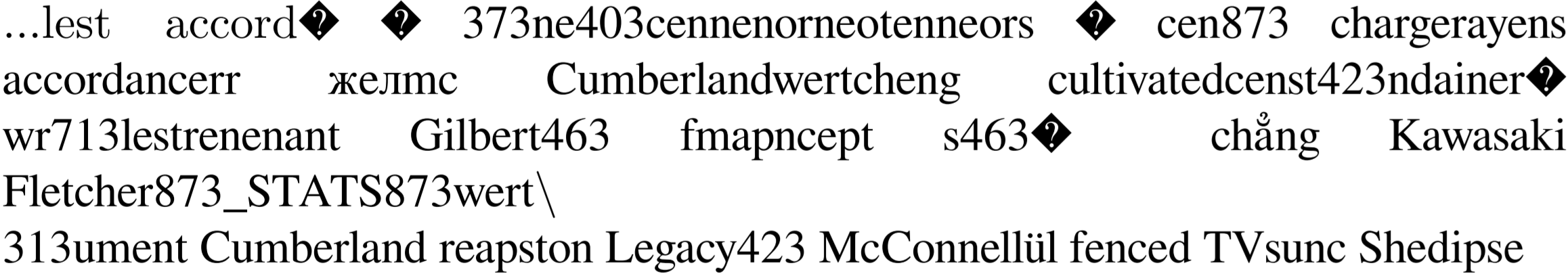}} \\
        ~ & 15276 & \raisebox{-0.85\totalheight}{\includegraphics[width=1\linewidth]{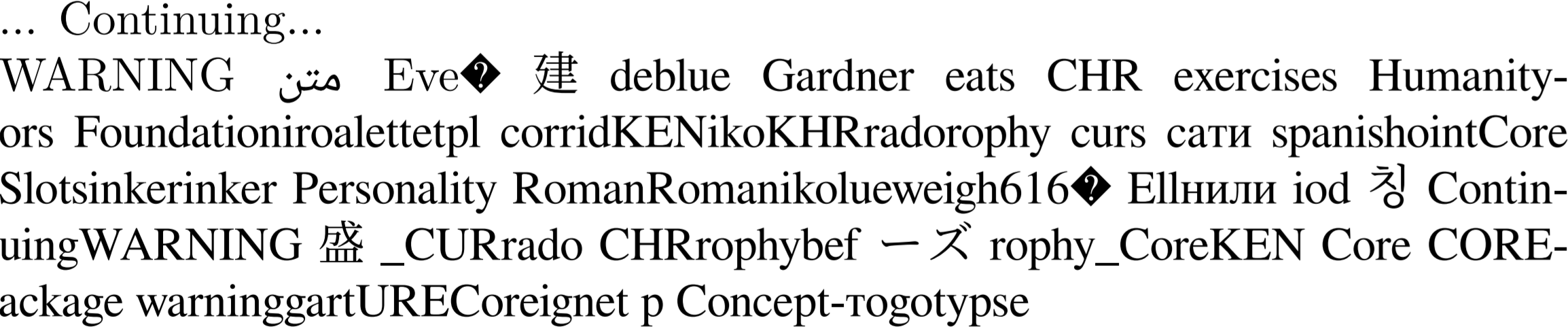}} \\
        ~ & 17350 & \raisebox{-0.85\totalheight}{\includegraphics[width=1\linewidth]{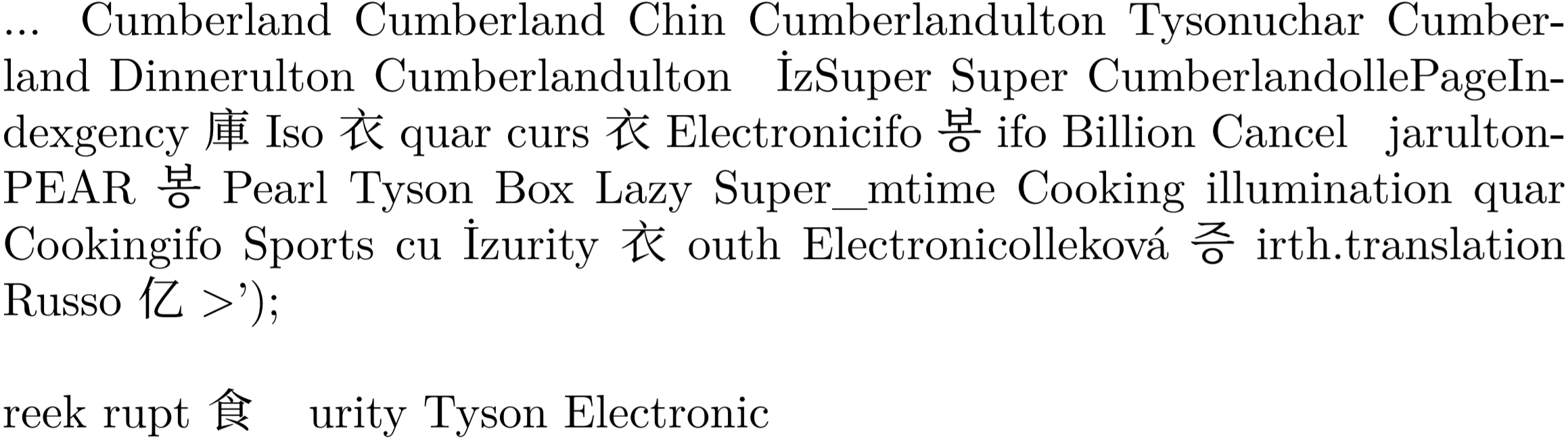}} \\
        ~ & 28590 & \raisebox{-0.85\totalheight}{\includegraphics[width=1\linewidth]{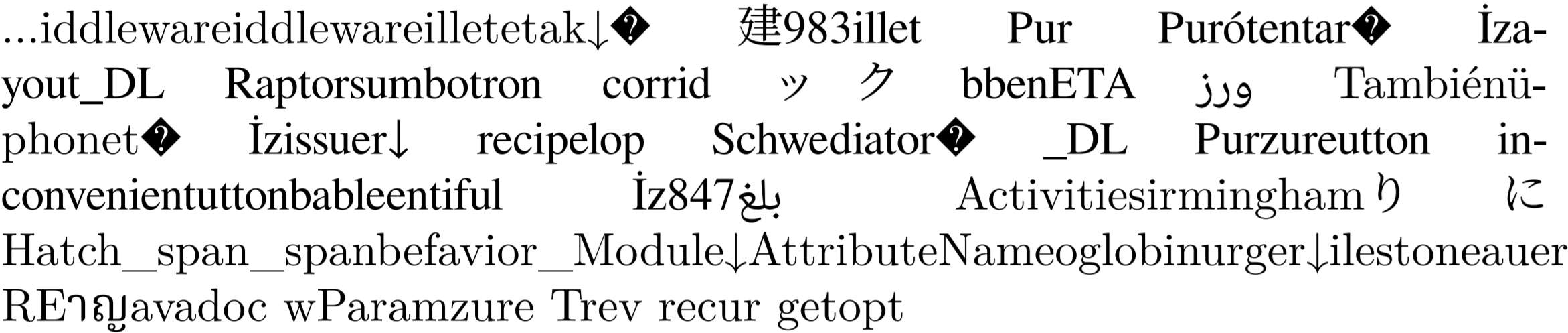}} \\ \hline
        \captionsetup{font=small}
        \caption{Selected negative steering strengths of all features containing the word \enquote{coffee} in layer 18. [Note that some outputs are displayed as pictures to ensure the correct rendering of uncommon characters.]}\label{tab:clamping_coffee_layer18_negative}
    \end{longtable}

\newpage
\subsection{Context-Dependence}\label{app:ContextDependence}
\subsubsection{\enquote{references to drinks, particularly wine and coffee} (18/7546)}\label{app:Context_CoffeeWine}
\begin{table}[ht!]
    \renewcommand{\arraystretch}{1.2}
    \small     
    \captionsetup{font=small}
    \centering
    \begin{tabular}{p{.15\textwidth} p{.25\textwidth} p{.5\textwidth}}
        \hline
        coffee  & cup \redhighlight{red!58.7}{\strut of} \redhighlight{red!48.55}{\strut coffee}    & hsd wekl dfsd coffee \\
        poison  & cup \redhighlight{red!58.7}{\strut of} \redhighlight{red!40.1}{\strut poison} & hsd wekl dfsd poison \\ 
        cup     & hot \redhighlight{red!21.5}{\strut cup} \redhighlight{red!80.68}{\strut of}   & We \redhighlight{red!77.29}{\strut drank} \redhighlight{red!100}{\strut coffee}\redhighlight{red!52.17}{\strut .} \\
        of      & hot \redhighlight{red!21.5}{\strut cup}   & We \redhighlight{red!77.29}{\strut drank} \redhighlight{red!56.52}{\strut poison}\redhighlight{red!52.42}{\strut .} \\
        hot     & hot \redhighlight{red!21.5}{\strut cup} \redhighlight{red!80.68}{\strut of} \redhighlight{red!61.84}{\strut coffee}   & This is the best museum I have ever\\
        cup \redhighlight{red!58.7}{\strut of}  & hot \redhighlight{red!21.5}{\strut cup} \redhighlight{red!80.68}{\strut of} \redhighlight{red!39,37}{\strut poison}   & Coffee\redhighlight{red!33.33}{\strut :} \redhighlight{red!25.12}{\strut This} \redhighlight{red!22.83}{\strut is} \redhighlight{red!23.07}{\strut the} \redhighlight{red!14.43}{\strut best} museum I have ever \\ \hline
    \end{tabular}
    \caption{Context-dependence of feature \enquote{references to drinks, particularly wine and coffee} (18/7546). The feature generally activates on tokens following the phrases \enquote{(hot) cup of} or \enquote{drank}. However, the features does not activate on the term \enquote{coffee} in isolation.}
    \label{tab:context_coffee_wine}
\end{table}

\subsubsection{\enquote{topics related to coffee, including its origins, effects, and cultural significance} (17/17320)}\label{app:Context_CoffeeOrigin}
\begin{table}[ht!]
    \renewcommand{\arraystretch}{1.2}
    \small     
    \captionsetup{font=small}
    \centering
    \begin{tabular}{p{.12\textwidth} p{.25\textwidth} p{.53\textwidth}} \hline
        \redhighlight{red!17.05}{\strut coffee}     & \redhighlight{red!17.05}{\strut coffee} \redhighlight{red!100}{\strut from} \redhighlight{red!59.47}{\strut Africa}   & kwentoeler welrl newr \redhighlight{red!15.81}{\strut caffeine} \\
        caffe\redhighlight{red!16.38}{\strut ine}   & \redhighlight{red!17.05}{\strut coffee} \redhighlight{red!100}{\strut from} \redhighlight{red!88.26}{\strut Ethiopia} & no sleep because the \redhighlight{red!11.36}{\strut caffeine} \\
        from                                        & beans \redhighlight{red!12.69}{\strut from} \redhighlight{red!57.95}{\strut Ethiopia}                                 &  This is the best museum I have ever\\
        Africa                                      & zebras from Africa                                                                                                    &  \redhighlight{red!23.39}{\strut Coffee}\redhighlight{red!74.24}{\strut :} \redhighlight{red!52.65}{\strut This} \redhighlight{red!39.02}{\strut is} \redhighlight{red!60.61}{\strut the} \redhighlight{red!47.35}{\strut best} museum I have ever \\
        Eth\redhighlight{red!7.77}{\strut i}\redhighlight{red!19.22}{\strut opia}   & :                       &  America's most popular equipment for \\
                                                    & \redhighlight{red!23.77}{\strut Coffee}\redhighlight{red!74.62}{\strut :}                                                         &  \redhighlight{red!23.39}{\strut Coffee}\redhighlight{red!74.24}{\strut :} \redhighlight{red!35.61}{\strut America}\redhighlight{red!66.67}{\strut 's} \redhighlight{red!83.71}{\strut most} \redhighlight{red!70.08}{\strut popular} \redhighlight{red!14.3}{\strut equipment} \redhighlight{red!71.59}{\strut for} \\ \hline
    \end{tabular}
    \caption{Context-dependence of feature \enquote{topics related to coffee, including its origins, effects, and cultural significance} (17/17320). The feature activates on tokens only in the surroundings of coffee-related terms.}
    \label{tab:context_coffeeOrigin}
\end{table}

\subsubsection{\enquote{mentions of coffee and related terms} (18/9463)}\label{app:Context_CoffeeRelated}
\begin{table}[ht!]
    \renewcommand{\arraystretch}{1.2}
    \small     
    \captionsetup{font=small}
    \centering
    \begin{tabular}{p{.15\textwidth} p{.27\textwidth} p{.48\textwidth}} \hline
        \redhighlight{red!93.61}{\strut coffee}         & \redhighlight{red!47.37}{\strut caffe}\redhighlight{red!35.71}{\strut ine}    & We drank our \redhighlight{red!71.05}{\strut coffee} \redhighlight{red!2.8}{\strut in} the morning \\
        \ \redhighlight{red!76.32}{\strut coffee}       & c\redhighlight{red!7.99}{\strut app}uch\redhighlight{red!47.74}{\strut ino}   & \redhighlight{red!0}{\strut This} is the best museum I have ever \\
        cup of \redhighlight{red!69.93}{\strut coffee}  & Es\redhighlight{red!57.52}{\strut presso} & \redhighlight{red!100}{\strut Coffee}\redhighlight{red!17,58}{\strut :} This is the best museum I have ever \\
        instant\redhighlight{red!47,74}{\strut coffee}  & C\redhighlight{red!27.63}{\strut af}\redhighlight{red!37.97}{\strut é}        & Roast (light, medium, dark) \\
        \redhighlight{red!93.61}{\strut coffee}break    & C\redhighlight{red!51.5}{\strut off}\redhighlight{red!5.87}{\strut in}        & They lowered the coffin \\
                                                        & Covf\redhighlight{red!21.34}{\strut efe}                                                    & njowdco f feekd\\
                                                        & iwern owk ail \redhighlight{red!89.47}{\strut coffee}                                                   & njowdco\redhighlight{red!71.05}{\strut ffe}\redhighlight{red!12.78}{\strut ek}d\\ \hline
    \end{tabular}
    \caption{Context-dependence of feature \enquote{mentions of coffee and related terms} (18/9463). The feature activates on \enquote{coffee} and related terms regardless of the context.}
    \label{tab:context_coffee}
\end{table}

\subsection{Curious Activations}\label{app:CuriousActivations}
\subsubsection{Steering: \enquote{monetary values and financial information} (20/6246, beginning of text activation: 156.0)}\label{app:CuriousActivation_monetaryvalues}
\begin{table}[ht!]  
    \small     
    \captionsetup{font=small} 
    \centering     
    \begin{tabular}{|p{.12\textwidth}|p{.82\textwidth}|}\hline

        Steering & The biggest event this week is... \\ \hline\hline
        Baseline & ... the Federal Reserve's Open Market Committee meeting, which begins on Tuesday. The Fed is widely expected to raise its target for the federal funds rate by \textbf{25 basis points} to a \textbf{range of 1.75\% to 2\%}. The Fed will also issue its \textbf{quarterly Summary of Economic Projections}, which includes forecasts for unemployment and inflation as well as projections [...] \\
        Coeff $-10$ & ... the Federal Open Market Committee meeting, and the Fed's decision on whether to \textbf{raise interest rates}. The FOMC meets \textbf{twice a year}, and this is one of those meetings. The Fed has been hinting that it will \textbf{raise interest rates}, but most analysts expect that it won't happen \textbf{this week}.\newline The FOMC meeting takes place on Tuesday [...]\\ \hline
    \end{tabular} 
    \caption{Steering capability of feature (20/6246), which is interpreted as \enquote{monetary values and financial information} and shows a beginning of text activation of $156.0$. Negatively clamping this feature suppresses to occurrence of explicitly financial terms (percentages, numbers and \enquote{quarterly}) to mentioning vague time specifications (e.g., \enquote{this week}) or terms without numbers (\enquote{raise interest rates}).} 
    \label{tab:clamping_-10_monetaryvalues} 
\end{table}

\subsubsection{Steering: \enquote{phrases related to weight loss and body shaping techniques} (20/21737, beginning of text activation: 100.5)}\label{app:CuriousActivation_weightloss}
\begin{table}[ht!]   
    \small     
    \captionsetup{font=small} 
    \centering     
    \begin{tabular}{|p{.12\textwidth}|p{.82\textwidth}|}\hline
        Steering & My goal for this year is... \\ \hline\hline
        Baseline & ... to \textbf{get back in shape} and to be able to \textbf{train more than just one day a week}. I have been training hard and eating well, but I still have a ways to go. My plan is to continue with my current program, but also add in \textbf{some new exercises} and some new techniques. I am hoping that by adding these new exercises I will [...] \\
        Coeff $-10$ & ... to \textbf{read 50 books} and I'm hoping to get there. I'm about a quarter of the way through my list and have \textbf{already read 13 books}. Some of them were short, but I've been trying to \textbf{read more than one book at a time} lately. I've also been listening to audiobooks, which has been helping me get [...] \\ \hline
    \end{tabular} 
    \caption{Steering capability of feature \enquote{phrases related to weight loss and body shaping techniques} (20/21737, beginning of text activation: 100.5). Negatively clamping this feature changes mentioned goals from exercise related to topics about reading.} 
    \label{tab:clamping_-10x_weightloss} 
\end{table}

\newpage
\subsection{Coherence in Steering}\label{app:CoherenceSteering}
\subsubsection{\enquote{references to the effects of coffee on health} (15/4922)}\label{app:CoherenceSteering_coffeehealth}
\begin{table}[ht!]        
    \small     
    \captionsetup{font=small}     
    \centering          
    \begin{tabular}{|p{.1\textwidth}|p{.84\textwidth}|}\hline
        Steering & My favorite drink is... \\ \hline\hline
        Baseline & ... \textbf{a margarita} but \textbf{my favorite cocktail is a mojito}. I have a lot of memories of enjoying mojitos with my friends and family.\newline I love to cook and bake, but I am not very good at it. My best dish is probably pasta carbonara. I make it every time I go to Italy.\newline My favorite thing to [...] \\
        Coeff $5$ & ... a \textbf{simple cup of coffee}. I love the flavor of \textbf{coffee} and the taste of \textbf{freshly ground beans}. I also love the aroma of freshly brewed coffee. I make sure to buy whole beans and grind them fresh for my first cup every day.\newline There are many ways to brew coffee, but there is nothing like a French press for making a delicious cup.\newline [...] \\
        Coeff $6$ & ... a \textbf{simple cup of hot tea}. I drink it without milk and sugar, and I like to enjoy the natural flavor of the \textbf{tea leaves}. I don't like to add any other flavors or sugar to it.\newline \textbf{I love coffee}, but if you want to make a good \textbf{cup of coffee}, you have to use \textbf{fresh beans} and grind it using a good [...] \\
        Coeff $8$ & ... a \textbf{simple cup of hot tea}, and I want to brew my favorite \textbf{cup of hot tea}. How do I make the best \textbf{cup of hot tea} that is brewed in my own favorite cup, or one of my favorite cups, for every tasting cup coffee grind grind coffee purchase grinding coffee grinder filter grinding groied grind coffee cup black cup he grog [...] \\ \hline
    \end{tabular} 
    \caption{Steering capability of feature \enquote{references to the effects of coffee on health} (15/4922). Positively clamping this feature by $5$ changes the output from alcoholic beverages to mentioning of \enquote{coffee}. Clamping it further ($6$, $8$) results in an increased generation of text on \enquote{tea}, before breaking down.} 
    \label{tab:clamping_5-6-8_coffeehealth} 
\end{table}

\newpage
\subsubsection{\enquote{references to pigeons and airborne topics} (18/16371)}\label{app:CoherenceSteering_pigeons}
\begin{table}[ht!]        
    \small     
    \captionsetup{font=small}     
    \centering          
    \begin{tabular}{|p{.1\textwidth}|p{.84\textwidth}|}\hline
        Steering & An example for a bird is... \\ \hline\hline
        Baseline & ... \textbf{the hummingbird}. \textbf{The hummingbird} is an insectivore and nectarivore. It feeds on insects and nectar from flowers. It has a long, slender bill that allows it to reach the nectar at the bottom of long, tubular flowers.\newline Another example for a bird is the penguin. The [...] \\
        Coeff $2$ & ... \textbf{the American Robin}. This bird can be seen in many different parts of the world. It is known for its beautiful red breast and its melodious song.\newline \textbf{The American Robin} \textbf{(Turdus migratorius)} is a migratory songbird that can be found in North America. It is a member of the thrush [...] \\
        Coeff $5$ & ... \textbf{the American Robin}. This bird can be seen in many different parts of the world. It is known for its beautiful red breast and has been used as a symbol of hope and renewal in many cultures. \textbf{The American Robin} is also known for its ability to fly long distances and has been used as a symbol of migration.\newline The [...] \\
        Coeff $10$ & ... \textbf{the American Robin}. This bird has an average length of 22.5 inches (57 cm) and its wingspan is 38.3 inches (97 cm). It weighs 16.4 ozs (465g) and has a record speed of 140 mph (224 km/hr). It has a lifespan [...] \\ \hline
    \end{tabular} 
    \caption{Steering capability of feature \enquote{references to pigeons and airborne topics} (18/16371). Despite its description, positively clamping this feature changes the mentioned bird from \enquote{hummingbird} to \enquote{American Robin}.} 
    \label{tab:clamping_2510_pigeon} 
\end{table}
\end{appendices}

\newpage 

\bibliography{bib}

@article{bricken2023monosemanticity,
       title={Towards Monosemanticity: Decomposing Language Models With Dictionary Learning},
       author={Bricken, Trenton and Templeton, Adly and Batson, Joshua and Chen, Brian and Jermyn, Adam and Conerly, Tom and Turner, Nick and Anil, Cem and Denison, Carson and Askell, Amanda and Lasenby, Robert and Wu, Yifan and Kravec, Shauna and Schiefer, Nicholas and Maxwell, Tim and Joseph, Nicholas and Hatfield-Dodds, Zac and Tamkin, Alex and Nguyen, Karina and McLean, Brayden and Burke, Josiah E and Hume, Tristan and Carter, Shan and Henighan, Tom and Olah, Christopher},
       year={2023},
       journal={Transformer Circuits Thread},
       url={https://transformer-circuits.pub/2023/monosemantic-features/index.html}
    }

@article{elhage2022superposition,
   title={Toy Models of Superposition},
   author={Elhage, Nelson and Hume, Tristan and Olsson, Catherine and Schiefer, Nicholas and Henighan, Tom and Kravec, Shauna and Hatfield-Dodds, Zac and Lasenby, Robert and Drain, Dawn and Chen, Carol and Grosse, Roger and McCandlish, Sam and Kaplan, Jared and Amodei, Dario and Wattenberg, Martin and Olah, Christopher},
   year={2022},
   journal={Transformer Circuits Thread},
   url={https://transformer-circuits.pub/2022/toy\_model/index.html}
}

@article{templeton2024scaling,
       title={Scaling Monosemanticity: Extracting Interpretable Features from Claude 3 Sonnet},
       author={Templeton, Adly and Conerly, Tom and Marcus, Jonathan and Lindsey, Jack and Bricken, Trenton and Chen, Brian and Pearce, Adam and Citro, Craig and Ameisen, Emmanuel and Jones, Andy and Cunningham, Hoagy and Turner, Nicholas L and McDougall, Callum and MacDiarmid, Monte and Freeman, C. Daniel and Sumers, Theodore R. and Rees, Edward and Batson, Joshua and Jermyn, Adam and Carter, Shan and Olah, Chris and Henighan, Tom},
       year={2024},
       journal={Transformer Circuits Thread},
       url={https://transformer-circuits.pub/2024/scaling-monosemanticity/index.html}
    }

@article{
bereska2024mechanistic,
title={Mechanistic Interpretability for {AI} Safety - A Review},
author={Leonard Bereska and Stratis Gavves},
journal={Transactions on Machine Learning Research},
issn={2835-8856},
year={2024},
url={https://openreview.net/forum?id=ePUVetPKu6},
note={Survey Certification, Expert Certification}
}

@INPROCEEDINGS{Rauker.2023,
author = {Räuker, Tilman and Ho, Anson and Casper, Stephen and Hadfield-Menell, Dylan },
booktitle = {2023 IEEE Conference on Secure and Trustworthy Machine Learning (SaTML) },
title = {{Toward Transparent AI: A Survey on Interpreting the Inner Structures of Deep Neural Networks }},
year = {2023},
volume = {},
ISSN = {},
pages = {464-483},
doi = {10.1109/SaTML54575.2023.00039},
url = {https://doi.ieeecomputersociety.org/10.1109/SaTML54575.2023.00039},
publisher = {IEEE Computer Society},
address = {Los Alamitos, CA, USA}
}

@misc{marks2025,
      title={Sparse Feature Circuits: Discovering and Editing Interpretable Causal Graphs in Language Models}, 
      author={Samuel Marks and Can Rager and Eric J. Michaud and Yonatan Belinkov and David Bau and Aaron Mueller},
      year={2025},
      eprint={2403.19647},
      archivePrefix={arXiv},
      primaryClass={cs.LG}
}

@incollection{Freiesleben.2026,
author="Freiesleben, Timo",
editor="Dur{\'a}n, Juan M.
and Pozzi, Giorgia",
title="Artificial Neural Nets and the Representation of Human Concepts",
bookTitle="Philosophy of Science for Machine Learning: Core Issues and New Perspectives",
year="2026",
publisher="Springer Nature Switzerland",
address="Cham",
pages="273--294",
isbn="978-3-032-03083-2",
doi="10.1007/978-3-032-03083-2_13"
}

@misc{Rai.2025,
      title={A Practical Review of Mechanistic Interpretability for Transformer-Based Language Models}, 
      author={Daking Rai and Yilun Zhou and Shi Feng and Abulhair Saparov and Ziyu Yao},
      year={2025},
      eprint={2407.02646},
      archivePrefix={arXiv},
      primaryClass={cs.AI}
}

@misc{Saphra.2024,
      title={Mechanistic?}, 
      author={Naomi Saphra and Sarah Wiegreffe},
      year={2024},
      eprint={2410.09087},
      archivePrefix={arXiv},
      primaryClass={cs.AI}
}

@misc{Bills.2023,
    title={Language models can explain neurons in language models},
    author={Bills, Steven and Cammarata, Nick and Mossing, Dan and Tillman, Henk and Gao, Leo and Goh, Gabriel and Sutskever, Ilya and Leike, Jan and Wu, Jeff and Saunders, William},
    year={2023},
    url = {https://openaipublic.blob.core.windows.net/neuron-explainer/paper/index.html}
      }

@misc{Durmus.2024,
author = {Esin Durmus and Alex Tamkin and Jack Clark and Jerry Wei and Jonathan Marcus and Joshua Batson and Kunal Handa and Liane Lovitt and Meg Tong and Miles McCain and Oliver Rausch and Saffron Huang and Sam Bowman and Stuart Ritchie and Tom Henighan and Deep Ganguli},
title = {Evaluating Feature Steering: A Case Study in Mitigating Social Biases},
date = {2024-10-25},
year = {2024},
url = {https://anthropic.com/research/evaluating-feature-steering},
}

@book{Elad.2010,
 author = {Elad, Michael},
 year = {2010},
 title = {Sparse and Redundant Representations},
 address = {New York, NY},
 publisher = {{Springer}},
 doi = {10.1007/978-1-4419-7011-4}
}

@online{Sharkey.2025,
      title={Open Problems in Mechanistic Interpretability}, 
      author={Lee Sharkey and Bilal Chughtai and Joshua Batson and Jack Lindsey and Jeff Wu and Lucius Bushnaq and Nicholas Goldowsky-Dill and Stefan Heimersheim and Alejandro Ortega and Joseph Bloom and Stella Biderman and Adria Garriga-Alonso and Arthur Conmy and Neel Nanda and Jessica Rumbelow and Martin Wattenberg and Nandi Schoots and Joseph Miller and Eric J. Michaud and Stephen Casper and Max Tegmark and William Saunders and David Bau and Eric Todd and Atticus Geiger and Mor Geva and Jesse Hoogland and Daniel Murfet and Tom McGrath},
      year={2025},
      eprint={2501.16496},
      archivePrefix={arXiv},
      primaryClass={cs.LG}
}

@online{NeurIPS.2025,
  author = {NeurIPS},
  title = {Mechanistic Interpretability Workshop},
  year = {2025},
  url = {https://mechinterpworkshop.com/},
  urldate = {2025-12-08}
}

@online{ICML.2024,
  author = {ICML},
  title = {Mechanistic Interpretability Workshop 2024},
  year = {2024},
  url = {https://icml2024mi.pages.dev/},
  urldate = {2025-12-08}
}

@article{Olah.2020,
  author = {Olah, Chris and Cammarata, Nick and Schubert, Ludwig and Goh, Gabriel and Petrov, Michael and Carter, Shan},
  title = {Zoom In: An Introduction to Circuits},
  journal = {Distill},
  year = {2020},
  url = {https://distill.pub/2020/circuits/zoom-in/},
  doi = {10.23915/distill.00024.001}
}

@article{London.2019,
  title={Artificial intelligence and black-box medical decisions: accuracy versus explainability},
  author={London, Alex John},
  journal={Hastings Center Report},
  volume={49},
  number={1},
  pages={15--21},
  year={2019},
  publisher={Wiley Online Library},
  doi = {10.1002/hast.973}
}

@misc{Chan.2023,
author = {Chan, Lawrence and Lang, Leon and Jenner, Erik},
title = {Natural Abstractions: Key Claims, Theorems, and Critiques},
date = {2023-03-16},
year = {2023},
url = {https://www.alignmentforum.org/posts/gvzW46Z3BsaZsLc25/natural-abstractions-key-claims-theorems-and-critiques},
}

@misc{Cunningham.2023,
      title={Sparse Autoencoders Find Highly Interpretable Features in Language Models}, 
      author={Cunningham, Hoagy and Ewart, Aidan and Riggs, Logan and Huben, Robert and Sharkey, Lee},
      year={2023},
      eprint={2309.08600},
      archivePrefix={arXiv},
      primaryClass={cs.LG}
}

@inproceedings{Mikolov.2013,
    title = "Linguistic Regularities in Continuous Space Word Representations",
    author = "Mikolov, Tomas  and
      Yih, Wen-tau  and
      Zweig, Geoffrey",
    editor = "Vanderwende, Lucy  and
      Daum{\'e} III, Hal  and
      Kirchhoff, Katrin",
    booktitle = "Proceedings of the 2013 Conference of the North {A}merican Chapter of the Association for Computational Linguistics: Human Language Technologies",
    month = jun,
    year = "2013",
    address = "Atlanta, Georgia",
    publisher = "Association for Computational Linguistics",
    url = "https://aclanthology.org/N13-1090/",
    pages = "746--751"
}

@online{Gao.2024,
      title={Scaling and evaluating sparse autoencoders}, 
      author={Leo Gao and Tom Dupré la Tour and Henk Tillman and Gabriel Goh and Rajan Troll and Alec Radford and Ilya Sutskever and Jan Leike and Jeffrey Wu},
      year={2024},
      eprint={2406.04093},
      archivePrefix={arXiv},
      primaryClass={cs.LG}
}

@online{Gao.2025,
      title={Weight-sparse transformers have interpretable circuits}, 
      author={Leo Gao and Achyuta Rajaram and Jacob Coxon and Soham V. Govande and Bowen Baker and Dan Mossing},
      year={2025},
      eprint={2511.13653},
      archivePrefix={arXiv},
      primaryClass={cs.LG}
}

@online{Arora.2018,
      title={Linear Algebraic Structure of Word Senses, with Applications to Polysemy}, 
      author={Sanjeev Arora and Yuanzhi Li and Yingyu Liang and Tengyu Ma and Andrej Risteski},
      year={2018},
      eprint={1601.03764},
      archivePrefix={arXiv},
      primaryClass={cs.CL}
}

@online{He_2024,
      title={Llama Scope: Extracting Millions of Features from Llama-3.1-8B with Sparse Autoencoders}, 
      author={Zhengfu He and Wentao Shu and Xuyang Ge and Lingjie Chen and Junxuan Wang and Yunhua Zhou and Frances Liu and Qipeng Guo and Xuanjing Huang and Zuxuan Wu and Yu-Gang Jiang and Xipeng Qiu},
      year={2024},
      eprint={2410.20526},
      archivePrefix={arXiv},
      primaryClass={cs.LG}
}

@misc{hijohnnylin_2025,
 author = {Lin, Johnny},
 editor = {GitHub},
 year = {2025},
 title = {automated-interpretability},
 url = {https://github.com/hijohnnylin/automated-interpretability},
 urldate = {2025-08-18}
}

@misc{OpenAI_2025b,
 author = {OpenAI},
 editor = {GitHub},
 year = {2025},
 title = {automated-interpretability},
 url = {https://github.com/openai/automated-interpretability},
 urldate = {2025-08-18}
}

@article{Geiger.2025,
  author  = {Atticus Geiger and Duligur Ibeling and Amir Zur and Maheep Chaudhary and Sonakshi Chauhan and Jing Huang and Aryaman Arora and Zhengxuan Wu and Noah Goodman and Christopher Potts and Thomas Icard},
  title   = {Causal Abstraction: A Theoretical Foundation for Mechanistic Interpretability},
  journal = {Journal of Machine Learning Research},
  year    = {2025},
  volume  = {26},
  number  = {83},
  pages   = {1--64},
  url     = {http://jmlr.org/papers/v26/23-0058.html}
}

@misc{Crosscoder.2024,
 author = {Lindsey, Jack and Templeton, Ady and Marcus, Jonathan and Conerly, Thomas and Batson, Joshua and Olah, Christopher},
 editor = {},
 year = {2024},
 title = {Sparse Crosscoders for Cross-Layer Features and Model Diffing},
 url = {https://transformer-circuits.pub/2024/crosscoders/index.html},
 urldate = {2025-09-15},
 date = {2024-10-25},
}

@misc{Neuronpedia_2023,
    title = {Neuronpedia: Interactive Reference and Tooling for Analyzing Neural Networks},
    year = {2023},
    url = {https://www.neuronpedia.org},
    author = {Lin, Johnny},
    urldate = {2025-10-07}
}

\end{document}